%% file: article.tex
\documentclass[9pt, journal]{IEEEtran}

\usepackage{cite}

\usepackage{amsmath,graphicx}
\usepackage{stmaryrd}
\usepackage{mathtools}
\usepackage{lmodern}
\usepackage{gensymb}
\usepackage{float}
\usepackage{placeins}
\usepackage{pgf}
\usepackage{multirow}
\usepackage{bigdelim}
\usepackage{makecell}
\usepackage{array}
\usepackage{arydshln}
\usepackage{amssymb}
\usepackage{bbm}
\usepackage{bm}

\usepackage{algorithm}
\usepackage{algorithmic}

\usepackage{pifont}

\usepackage{enumitem}

\newcommand{\tablecbottom}[1]{\Xcline{#1}{0.35ex}}
\newcommand{\tablectop}[1]{\Xcline{#1}{0.35ex}}

\newcommand{\tablebottom}{\noalign{\hrule height 0.35ex}}
\newcommand{\tabletop}{\noalign{\hrule height 0.35ex}} %

\def\lblsec{Section } %
\def\lblsecs{Sections } %
\def\lbltab{Table } %
\def\lbleq{Equation } %
\def\lbleqabbr{Eq.} %
\def\lblfig{Fig. } %
\def\lblalg{Alg. } %

\def\x{{\mathbf x}}
\def\z{{\mathbf z}}
\def\R{{\mathbb R}}
\def\normset{\mathcal{X}_{\text{norm}}}
\def\trainset{\mathcal{X}_{\text{train}}}
\def\softmax{{\operatorname{smax}}}

\def\bsize{\alpha}
\def\Ipart{I_{\text{part}}}
\def\nrs{n_{\text{tsp}}}
\def\nrstest{n_{\text{sp}}}

\def\loss{\mathcal{L}}
\def\crossentropy{\loss_{\text{CE}}}
\def\lossPzl{\loss_{\text{pzl}}}
\def\lossIp{\loss_{\text{piece}}}
\def\lossTint{\loss_{\text{tint}}}
\def\lossCol{\loss_{\text{col}}}

\def\lossupbranch{\loss_{\text{U-branch}}}
\def\losslowbranch{\loss_{\text{L-branch}}}

\newcommand{\datagen}[1]{DG_{#1}}

\def\inImg{I}
\newcommand{\inImgPiece}[1]{R_{#1}}

\newcommand{\floor}[1]{\lfloor #1 \rfloor}

\newcommand{\pixclstr}[3]{{#1_{#2}^{(#3)}}}
\newcommand{\expect}[2][]{\operatorname{E}_{#1}\left[#2\right]}
\newcommand{\normdist}[3]{\mathcal{N}\left(#1;#2,#3\right)}

\renewcommand{\checkmark}{\ding{51}}%
\newcommand{\crossmark}{-}%

\def\captionfont{\normalfont\sffamily\footnotesize}

\newcommand{\groupcell}[4]{\multirow{#1}{*}{}\ldelim\{{#1}{#2}[\parbox{#3}{#4}\ ]}

\begin{document}
\bstctlcite{BSTnolink}
\title{Efficient Anomaly Detection Using Self-Supervised Multi-Cue Tasks}

\author{Loïc~Jézéquel,
	    Ngoc-Son~Vu,
	    Jean~Beaudet,
	    and~Aymeric~Histace %
}

\maketitle

\begin{abstract}
	Anomaly detection is important in many real-life applications. Recently, self-supervised learning has greatly helped deep anomaly detection by recognizing several geometric transformations. However these methods lack finer features, usually highly depend on the anomaly type, and do not perform well on fine-grained problems. To address these issues, we first introduce in this work three novel and efficient discriminative and generative tasks which have complementary strength: (i) a piece-wise jigsaw puzzle task focuses on structure cues; (ii) a tint rotation recognition is used within each piece, taking into account the colorimetry information; (iii) and a partial re-colorization task considers the image texture. In order to make the re-colorization task more object-oriented than background-oriented, we propose to include the contextual color information of the image border via an attention mechanism. We then present a new out-of-distribution detection function and highlight its better stability compared to existing methods. Along with it, we also experiment different score fusion functions. Finally, we evaluate our method on an extensive protocol composed of various anomaly types, from object anomalies, style anomalies with fine-grained classification to local anomalies with face anti-spoofing datasets. Our model significantly outperforms state-of-the-art with up to 36\% relative error improvement on object anomalies and 40\% on face anti-spoofing problems.
\end{abstract}


\IEEEpeerreviewmaketitle

\section{Introduction}
\label{sec:intro}

\IEEEPARstart{O}{ne} of the most fundamental challenge in machine learning is detecting an observation as anomalous compared to a normal baseline. Properly solving such problem with high predictability and robustness has been essential in many fields. To mention a few, in intrusion detection \cite{netintrusion2019} where we wish to detect untrustworthy entries on a network, fraud detection \cite{onlinefraud2018} where a forged item or transaction must be rejected, in medical imaging \cite{DBLP:journals/tip/KumarA20} where abnormalities in a captured image must be located, video surveillance \cite{DBLP:journals/tip/LvZCXLY21,DBLP:journals/tip/LeyvaSL17} where abnormal events are detected, and in manufacturing defect detection \cite{DBLP:journals/tip/ZengLFC21,DBLP:journals/tip/ZouZLQWW19}.

With the advent of deep learning, many tasks on image data including binary classification and anomaly detection (AD) have greatly improved. Nevertheless classical binary classification still generally lacks robustness and reliability outside its training domain. Many anomaly detection methods try to solve this problem by only learning the normal class boundary, rather than directly discriminating anomalies from normal samples. Any observation defined outside is then deemed as anomalous. This decision rule is especially useful when the anomaly class boundary is ill-defined or continually evolving and only few anomalous training samples are available.

The recent explosion of self supervision further improves unsupervised learning abilities and reduces the needed amount of labeled data. It enables to discriminate anomalies from normal samples by learning to solve simple tasks such as geometric transformation classification. However, although deep anomaly detection can achieve interesting performance, it still suffers from limitations on more challenging problems with local and fine-grained differences between anomalies and normal samples. Indeed, existing self-supervised anomaly detection algorithms evaluated their performance on datasets like CIFAR10 or CIFAR100 but not on fine-grained ones like Caltech-Birds or face anti-spoofing. Moreover, these methods usually have an high inference time, making them impractical for real-life anomaly detection problems. For example, the state-of-the-art model GeoTrans \cite{Golan2018DeepAD} needs to apply during inference 72 different transformations to the input making it around 10 times slower than our proposed method.

In this given context, our main contributions in this paper are the following:
\begin{itemize}
    \item We introduce a new way to efficiently exploit the benefits of discriminative and generative auxiliary tasks in self-supervised anomaly detection. Using the two-branch network, we are among the first to reach high-quality results with auxiliary tasks on fine-grained anomaly detection and face anti-spoofing in a one-class setting.
    \item We carefully design and optimize three novel specialized auxiliary tasks according to loss functions, anomaly scores as well as complexity. This allows our model to learn very rich and complementary representations which better encompass image structure (\lblsec \ref{subsec:pw-puzzle}), colorimetry (\lblsec \ref{subsec:tint-rot}) and texture (\lblsec \ref{subsec:part-color}). With these tasks, we also explore different out-of-distribution (OOD) detection methods and fusion functions.
	\item We compare our method with state-of-the-art using an exhaustive protocol for anomaly detection covering object, style and local anomalies, and even more challenging task of face anti-spoofing.
	\item The proposed method obtains high-quality results with up to 36\% AUROC relative improvement on object anomalies and 53\% on face anti-spoofing from state-of-the-art anomaly detection methods.
\end{itemize}

This paper follows the motivation of our work presented in \cite{Jezequel2021}. In \cite{Jezequel2021}, we improved the anomaly detection by simultaneously solving in a self-supervised fashion a high-scale geometric task and a low-scale jigsaw puzzle task. It is worth noting that the differences of this paper compared to \cite{Jezequel2021} are significant: all pretext tasks are novel and more efficient. In this paper, we address the inference complexity issue and considerably improve the anomaly detection performance.

First, we give an overview of anomaly detection related work in \lblsec\ref{sec:related}. Then we present our new pretext tasks in \lblsec\ref{sec:pretext-tasks}, and our study of OOD methods with fusion in \lblsec\ref{sec:ood-methods}. Our complete model is summarized in \lblsec\ref{sec:method} which we give a general overview in \lblfig \ref{fig:method-overview}. In a first stage, a jigsaw puzzle task with intra-piece tint rotation detection and a partial colorization are performed. Then in a second stage, a set of OOD scores is computed for each task and is aggregated into a single anomaly score using a fusion function. In addition, we extensively compare our model with state-of-the-art in \lblsec\ref{sec:results}, and provide several experiments on the influence of our model parameters in \lblsec\ref{sec:param-study}. Finally, we discuss future work in \lblsec\ref{sec:conclusion}.

\section{Related work}
\label{sec:related}

We first review several common classical and deep anomaly detection methods in \lblsec\ref{sec:related}.A and \lblsec\ref{sec:related}.B. We then present self-supervised learning and how they are applied for AD in \lblsec\ref{sec:related}.C and \lblsec\ref{sec:related}.D, respectively. Readers are refereed to \cite{Ruff_2021,Salehi2022Openset,Jaiswal2020SSLSurvey} for more in-depth surveys on AD or self-supervised learning.

\subsection{Classical anomaly detection}

The main goal in anomaly detection is to classify a sample as normal or anomalous. Formally, we predict $P(\x\in\normset)$ for an observation $\x$ and a normal (or positive) class $\normset$. The anomalous (or negative) class is then defined implicitly as the complementary of the normal class in image space.
We can generally categorize anomalies into three families:

\begin{enumerate}
	\item \textbf{Object anomaly}: any object which is not included in the positive class, e.g., a cat is an object anomaly in regards to dogs.
	\item \textbf{Style anomaly}: observations representing the same object as the positive class but with a different style or support, e.g., a realistic mask or a printed face represent faces but with a visible different style.
	\item \textbf{Local anomaly}: observations representing and sharing the same style as the positive class, however a localized part of the image is different. Most of the time, these anomalies are the superposition of two generative processes, e.g., a fake nose on a real face is a local anomaly.
\end{enumerate}

\begin{figure}[htb]
	\centering
	\begin{minipage}[b]{0.3\linewidth}
		\centering
		\centerline{\includegraphics[width=\linewidth]{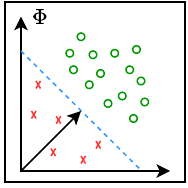}}
		\vspace{.1cm}
		\centerline{\captionfont (a) OC-SVM}
	\end{minipage}
	\hfill
	\begin{minipage}[b]{0.3\linewidth}
		\centering
		\centerline{\includegraphics[width=\linewidth]{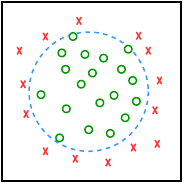}}
		\vspace{.1cm}
		\centerline{\captionfont (b) SVDD}
	\end{minipage}
	\hfill
	\begin{minipage}[b]{0.3\linewidth}
		\centerline{\includegraphics[width=\linewidth]{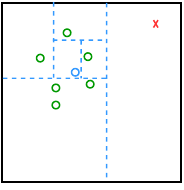}}
		\vspace{.1cm}
		\centerline{\captionfont (c) IF}
	\end{minipage}
	\caption{Overview of classical methods where green circles are normal samples and red cross anomalies. In (a) and (b) the anomalies are not part of the training dataset. In (c) the sample on the right is predicted as anomalous since it only required a single partition, while the blue circle is deemed normal.}
	\label{fig:classical-ad}
\end{figure}

Usually, we assume in anomaly detection that only normal samples are available during training, meaning that methods are in one-class setting. Traditionally, one-class Support Vector Machine \cite{Schlkopf1999SupportVM} (\textbf{OC-SVM}) or its extension the Support Vector Data Description \cite{Tax2004} (\textbf{SVDD}) were used for anomaly detection. The anomaly score of an observation $\x$ is given by its distance to a parameterized boundary $\Omega$. OC-SVM defines $\Omega$ as an hyper-plan separating the origin from the normal samples with the maximum margin, whereas SVDD uses an hyper-sphere containing all normal samples with the minimum radius (see \lblfig \ref{fig:classical-ad}(a,b)). 

Fully-unsupervised methods which learn from a set of unlabeled data containing normal samples and anomalies were also used. Such non-deep methods include Robust Principal Component Analysis \cite{10.1145/1970392.1970395} (\textbf{RPCA}) or the Isolation Forest (\textbf{IF}) \cite{isolationforest2009}. Rather than modeling the normal samples, the IF algorithm tries to isolate anomalies from normal samples via successive random partitions of the feature space. If the sample can be entirely isolated (i.e. be the only point in a region) in a few partitions, then it is more likely to be anomalous (see \lblfig \ref{fig:classical-ad}(c)).

These classical methods have shown great success on low-dimensional data such as tabular data, but usually fail on higher dimension inputs such as images.

\subsection{Deep anomaly detection}

The introduction of neural networks as feature extractors gave birth to several hybrid methods where a pre-trained neural network is used to extract features, on which a classical algorithm such as OC-SVM or isolation forest is trained. It ultimately led to the first end-to-end anomaly detection neural network, the one-class Neural Network (\textbf{OC-NN}) \cite{Chalapathy2019} which integrates the OC-SVM loss in the network training. More recent methods include different dedicated approach to anomaly detection. In \cite{Ngo2019,Oza2019,goyal2020DROCC}, a binary classification is used with pseudo negative images or latent vectors to represent the anomaly class. Another approach is to use the error of a generative model reconstruction \cite{schlegl2017anogan,akcay2018ganomaly, Tuluptceva2019PerceptualIA, DBLP:journals/corr/abs-2109-14020} or the gradient of the error given that the image is normal \cite{Kwon2020}. Finally, the self-supervision framework can be used to learn normal class representations and subsequently form an anomaly score  as presented in \lblsec \ref{subsec:sslad}.

There also have been semi-supervised anomaly detection methods such as DeepSAD \cite{deepsad2020} or deviation networks \cite{Pang2019DeepAD} where we assume some of the anomalies representing a few modes are available. These methods can achieve better accuracy on borderline cases given enough diverse anomalies, which is often less manageable in practice. In particular, these two methods directly learn representations by minimizing the distance of normal sample features to an hypersphere center, while maximizing the distance to the anomalies. It follows the compactness principle, where the normal class representations variance is minimized and the inter-class representations variance is maximized. 

\subsection{Self-supervised learning}

Self supervised learning (SSL) is a part of representation learning, where useful and general representations are learned from an unlabeled dataset. The learned features are then used through transfer learning for a different task such as classification.

In this manner, representations are learned by solving from the data an auxiliary task $\mathcal{T}$, which is often unrelated to the final one. The pretext task can either be discriminative, usually resulting in a multi-class classification setting or generative where a regression loss is often utilized. Any SSL is defined by its \textit{pretext objective loss} $\loss$ and its \textit{pretext data generation function} $\datagen{\mathcal{T}}:\mathcal{P}(\mathcal{X})\mapsto\mathcal{P}(\mathcal{X}\times K)$ which yields a labeled set from an unlabeled set $\mathcal{X}$. In the case of discriminative tasks, it is usually done via $n$ images transformations $T_1,\cdots,T_n$:
\begin{equation}
	\datagen{\mathcal{T}}\left(\{\x_i\}_{i\in\llbracket 1,N \rrbracket}\right)=\{(T_j(\x_i), j)\}_{i\in\llbracket 1,N \rrbracket,j\in\llbracket 1,n \rrbracket}
\end{equation} where the $\x_i$ are images from the unlabeled training dataset.

In other words, SSL consists of two steps: \textbf{(1)} generating a labeled set $\mathcal{X_T}=\datagen{\mathcal{T}}(\mathcal{X})$, \textbf{(2)} training a classification or regression network on this generated labeled set. One of the final layers are thus used as a feature extractor. Some commonly used tasks are: 90° rotation prediction \cite{Gidaris2018UnsupervisedRL}, jigsaw puzzle \cite{jigsawssl2016}, distortions \cite{examplar2014}, colorization \cite{Zhang2016ColorfulIC}, image inpainting \cite{Pathak2016ContextEF} or relative patches prediction \cite{patchpos2015}. 

More recently, the contrastive learning framework \cite{LeKhac2020} has been extensively used for self-supervised representation learning. Unlike the methods above, it does not rely on an explicit pretext task and directly formulates losses on the representations. The most effective contrastive method is instance discrimination \cite{Chen2020,Grill2020} where the objective is to maximize similarity between augmented versions of a same image (positive samples) while minimizing similarity with any other images (negative samples). The instance discrimination can be seen as a pretext task where the pretext data generation function maps samples to the set of positive pairs and negative pairs and the objective function is to discriminate positive from negative pairs using cosine similarity in representation space.

\subsection{SSL anomaly detection}
\label{subsec:sslad}

In this section, we first present how to apply SSL for AD and then discuss some state-of-the-art methods exploiting SSL for AD.

Very recently, SSL has been adapted to the one-class anomaly detection framework. First we learn to solve an auxiliary task in a SSL fashion. Then, a measure of how well the network can solve the task on the generated dataset $\datagen{\mathcal{T}}(\mathcal{X})$ is used to classify at inference time an observation $\x$ as anomalous or normal. The main assumption is that the network will perform relatively well on normal samples but will fail on anomalies. The goals of representation learning and AD are different. In representation learning we try to maximize the performance of the representation on as many downstream tasks and data as possible; whereas in AD, we want a clear discrimination through performance on normal and anomalous data. 

\begin{figure}[tbh]
	\centering
	\includegraphics[width=0.8\linewidth]{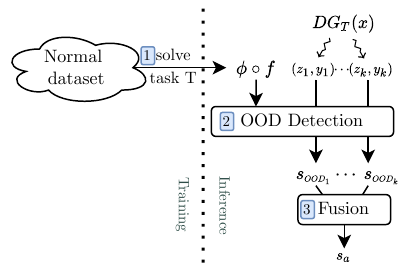}
	\caption{The three steps of pretext task based self-supervised learning anomaly detection: (1) the pretext task is solved on the normal dataset, (2) OOD detection functions are applied during inference on a pretext dataset generated via the data generative function on the unseen sample, and (3) these OOD scores are aggregated into a single anomaly score.}
	\label{fig:ssl-ad}
\end{figure}

Any SSL anomaly detector is composed of three steps (see \lblfig\ref{fig:ssl-ad}):

\begin{enumerate}
	\item The \textbf{representation learning} on the normal class, carried out in a self-supervised manner. In our case this is done by solving a pretext task $\mathcal{T}$, but other methods employ other mechanisms such as contrastive learning.
	
	\item During inference of an unseen sample $\x$, an \textbf{out-of-distribution (OOD) detection method} is applied on the generated labeled samples $\datagen{\mathcal{T}}(\{\x\})$. The goal of OOD methods is to detect whether or not an observation has been sampled from the same distribution as the training set. OOD is more low-level and general than AD, and aims at modeling the training distribution rather than the normal class. For example, contrary to AD the CIFAR-100 dataset would be considered out of distribution in regards to CIFAR-10. Given a pre-trained model $\Psi$ on a distribution $F_{\trainset}$, it estimates $P(\x\sim F_{\trainset})$. The normal training set is assumed to be close enough to the real distribution of normal samples, and since we have access to the correct task label $y$, the following approximations hold: \begin{equation}
		s_{OOD}((\x, y);\Psi)\approx P(\x\sim F_{\trainset})\approx P(\x\in\normset)
	\end{equation} where $s_{OOD}((\x, y);\Psi)$ is the OOD score for an image $\x$ with its label $y$ given the pre-trained network $\Psi$.
	
	\item The \textbf{fusion of the OOD scores} into a single anomaly score $s_a$ using a fusion function $M$.
\end{enumerate}

In the rest of this section, we detail several state-of-the-art self-supervised anomaly detection algorithms that are the most closely related to our work.

In \textbf{GeoTrans} \cite{Golan2018DeepAD}, the auxiliary task is to classify which geometrical transformation has been applied to the input from a set $\{T_i\}$ of 72 random composition of translations, rotations and symmetries. At the end of training, a Dirichlet distribution parameterized by $\tilde{\boldsymbol{\alpha}}_{i}$ is fitted over the softmax responses of each transformation on the normal class $\mathbf{y}\left(T_{i}(\x)\right)=\softmax(\phi\circ f(\x))$; then its log-likelihood is used during inference.

\begin{equation}
	s_a(\x)=\sum_{i=1}^{72}\left(\tilde{\boldsymbol{\alpha}}_{i}-1\right) \cdot \log \mathbf{y}\left(T_{i}(\x)\right)
\end{equation}

In \textbf{MHRot} \cite{Hendrycks2019UsingSL}, the task is to simultaneously classify 90° rotations, horizontal translations ($\operatorname{VTrans}$), and vertical translation ($\operatorname{HTrans}$), each modeled by a softmax head. Accordingly, the pretext data generation function is the composition $T_{r,s,t}=\operatorname{Rot}(r)\circ \operatorname{HTrans}(s)\circ \operatorname{VTrans}(t)$, where $r\in\{0\degree,90\degree,180\degree,270\degree\}$, $s\in\{0,-t_x,+t_x\}$ and $t\in\{0,-t_y,+t_y\}$. During inference, the three softmax of the known transformations for each of the 36 transformation compositions are summed as anomaly score:

\begin{equation}
	s_a(\x)=\sum_r \sum_s \sum_t \quad \mathbf{y}(T_{\scriptscriptstyle r,s,t}(\x))_{r,s,t}
\end{equation}

Another class of models, called \textbf{two-stage anomaly detectors} \cite{DBLP:conf/iclr/SohnLYJP21}, does not use the representation learning task during inference, but rather directly apply OOD methods on the representation space \cite{tack2020csi,sehwag2021ssd,DBLP:journals/corr/abs-2106-03844,DBLP:journals/corr/abs-2103-15296}.
For example, in \textbf{SSD} \cite{sehwag2021ssd} the representation learning step is performed through contrastive learning, then OOD detection is applied on the representation space induced by the encoder $\phi$. The training data representations are clustered around several centroids using K-means. The Mahalanobis distance is used to compute the anomaly score:

\begin{equation}
	s_a(\x)=\min_m (\phi(\x)-\mu_m)^T \Sigma_m^{-1} (\phi(\x)-\mu_m)
\end{equation} Similarly, \textbf{DROC-contrastive} (Deep Representation One-class Classification) \cite{DBLP:conf/iclr/SohnLYJP21} first learn self-supervised representations from one-class data, and then build one-class classifiers on learned representations. Contrastive learning with distribution augmentation is used for the self-supervised representation learning, and a OC-SVM for the one-class classification.

Finally, it is interesting noting that some SSL anomaly detectors solve the more specific task of anomaly segmentation like CutPaste \cite{DBLP:conf/cvpr/LiSYP21}, SOMAD \cite{DBLP:conf/icip/LiJMWHG21}. Those anomaly segmentation consists in predicting a heatmap where the anomaly score is computed on each pixels of the input image. They usually consider very minute and local AD, such as defect detection, while in this work we focus on image-level anomaly detection.

\begin{figure*}[t]
	\centering
	\includegraphics[width=0.95\linewidth]{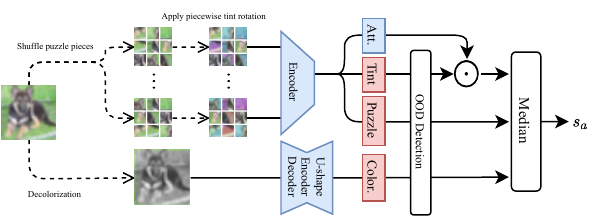}
	\caption{Method overview. Our model consists of discriminative (upper U-branch) and generative (lower L-branch) tasks. All the discriminative tasks share the same encoder.}
	\label{fig:method-overview}
\end{figure*}

\newcommand{\tbeqref}[1]{\hfill{\tiny (\lbleqabbr\ref{#1})}}

\begin{table*}[!htbp]
	\caption{\label{tab:task-overview}Overview of the loss function and OOD score for each proposed task. Upper U-branch consists of piece-wise puzzle, tint rotation tasks while lower L-branch consists of partial colorization task.}
	\centering
	\resizebox{1 \textwidth}{!} {
	\begin{tabular}{l|ll}
		\tabletop
		\textbf{Task (type)}  & \textbf{Loss} & \textbf{Anomaly score} \\ 
		\hline
		\makecell[cl]{Piece-wise puzzle \\ {\scriptsize(\textit{Cross-entropy})}}& $\lossPzl(\inImg)\propto\sum\limits_{i=1}^n \crossentropy(\phi\circ f_i(I); \Pi_i)$ \tbeqref{eqn:puzzle-loss}
		         & $s(\inImg)=\frac{1}{n}\sum\limits_{i=1}^{n} s_{OOD}((\Pi(\inImg), \Pi_i), \phi\circ f_i)$ \tbeqref{eqn:puzzle-score}      \\
		\makecell[cl]{Tint rotation \\ {\scriptsize(\textit{Expected L1 error})}} & $\lossTint(\inImg)\propto\expect[\Theta\sim\phi\circ f(\inImg)]{\|\inImg-\gamma(\inImg,\theta-\Theta)\|_1}$ \tbeqref{eqn:tint-loss}  & $s(\gamma(\inImg,\theta))=\sum\limits_{i=1}^c \softmax{(\phi\circ f(\gamma(\inImg, \theta)))}_i \left(\frac{\|\inImg-\gamma(\inImg,\theta-i\cdot\frac{2\pi}{c})\|_1}{W\times H\times 255}\right)$ \tbeqref{eqn:tint-score} \\
		\hdashline
		\makecell[cl]{Partial colorization \\ {\scriptsize(\textit{Expectation Max.})}} & $\lossCol(\inImg)=\sum\limits_{ij}\sum\limits_{k=1}^{K}Q_{\text{EM}}\left(\pixclstr{\pi}{ij}{k}, \pixclstr{\mu}{ij}{k}, \pixclstr{\Sigma}{ij}{k}\right)$ \tbeqref{eqn:color-loss} &  $s(A_{ij},B_{ij}|\Ipart)=\sum\limits_{k=1}^K \pixclstr{\pi}{ij}{k} \normdist{A_{ij},B_{ij}}{\pixclstr{\mu}{ij}{k}}{\pixclstr{\Sigma}{ij}{k}}$ \tbeqref{eqn:color-score} \\
		\tablebottom
	\end{tabular}
	}
\end{table*}

\section{Novel pretext tasks}
\label{sec:pretext-tasks}

In the rest of the paper, we consider an observation $z$, its label $y$ and a pre-trained network $\phi\circ f$. We gradually detail the proposed pretext tasks for anomaly detection which focus on different visual cues: structure, colorimetry and texture. The tasks of piece-wise puzzle, tint rotation and their combination are discriminative (\lblsecs\ref{subsec:pw-puzzle}, \ref{subsec:tint-rot}, \ref{subsec:intra-piece-tasks}) whereas the colorization task is generative (\lblsec\ref{subsec:part-color}). An overview of the loss function and anomaly score for each proposed task is shown in \lbltab\ref{tab:task-overview}.

\subsection{Piece-wise puzzle task}
\label{subsec:pw-puzzle}

The puzzle task has been successfully used as a pretext task for representation learning \cite{jigsawssl2016,Carlucci2019}. First an image is separated into $n=n_w\times n_h$ pieces, with some random margin between them. Then given the an image generated by shuffling pieces, a deep encoder is trained to predict which permutation has been applied. It is therefore formulated as a classification task where the prediction label corresponds to the index of the permutation among the $n!$ total possibilities. When the number of pieces becomes too large, the full task is not conceivable and the model should only learn to classify a smaller random subset of all permutations. This formulation of the jigsaw puzzle task, used in our previous work \cite{Jezequel2021} along with geometrical transformation recognition, enables our model to learn low-scale fine features. In the rest of the paper, we call this formulation the \textit{partial puzzle task}. It is worth noting that regarding to our previous work \cite{Jezequel2021}, this paper reconsiders only the puzzle task which is further optimized in both term of time and performance, as will be described in the rest of this section, while \textit{other tasks including tint rotation and partial colorization have never been used} for visual anomaly detection in the literature, to the best of our knowledge.

The partial puzzle task \cite{Jezequel2021} has several limitations: (i) the quality of the representation highly depends on the chosen permutations. Indeed if the sampled permutations are too hard (e.g. swapping two corners) or too easy, the learned representations will suffer; (ii) Moreover from an anomaly detection perspective, all mispredicted permutations are equally penalized regardless of the number of misplaced pieces.

To address these limitations, we propose here an improved piece-wise puzzle task. Rather than predicting the permutation index, we train a deep encoder to predict the original position of each piece. By assuming each piece is independent, we can now cover all the permutations with only $n^2$ outputs instead of $n!$. Thereby we separate the output layer $f$ into $n$ functions $f_1,\cdots,f_n$, each corresponding to a piece. 

\begin{figure}[tbh]
	\centering
	\includegraphics[width=1\linewidth]{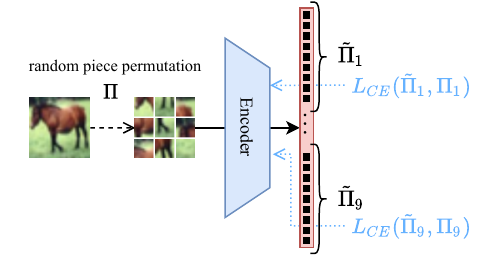}
	\caption{Piece-wise puzzle task for $3\times 3$ pieces, where $\Pi$ is a random piece permutation and $\tilde{\Pi}_i$ is the prediction vector for the j\textsuperscript{th} piece (\lblsec \ref{subsec:pw-puzzle}).}
	\label{fig:pw-puzzle}
\end{figure}

Let $\Pi$ be a random permutation, $\Pi(\inImg)$ corresponds to the image $I$ where each piece has been moved according to $\Pi$, and $\Pi_i$ corresponds to the new position of the i\textsuperscript{th} piece. The task is learned using the cross-entropy loss $\crossentropy$ on every piece predictions:
\begin{equation}
	\label{eqn:puzzle-loss}
	\lossPzl(\Pi(I))=\frac{1}{n}\sum_{i=1}^n \crossentropy(\phi\circ f_i(\Pi(I)); \Pi_i)
\end{equation} The full task is illustrated in \lblfig \ref{fig:pw-puzzle}.
In practice, we sample during every training epoch a random subset of $\nrs$ permutations for each normal image. In order to have as many different permutations as possible in the training set, we define $\nrs=\frac{n!}{N_{train}\cdot ep}$, where $N_{train}$ is the size of the training set and $ep$ the number of training epochs.

During inference we also consider a random subset of $\nrstest$ permutation, and compute an anomaly score for each of them:
\begin{equation}
\label{eqn:puzzle-score}
s_a(\Pi(\inImg))=\frac{1}{n}\sum_{i=1}^{n} s_{OOD}((\Pi(\inImg), \Pi_i), \phi\circ f_i)
\end{equation} where $s_{OOD}$ is an OOD score function which is presented in more detail in \lblsec\ref{sec:ood-methods}. In fact, we try different OOD functions and find out the best one. While $\nrs$ permutations are randomly used during training, it is important to note that the $\nrstest$ permutations are fixed for all tests in the final model.

With this new piece-wise puzzle task, lower anomaly detection errors can be reached while keeping the same inference complexity as the partial puzzle task (see results in \lblfig\ref{fig:puzzle-comp}).

\subsection{Tint rotation task}
\label{subsec:tint-rot}

High-scale object colorimetry is a simple but powerful clue to discriminate anomalies, especially in spoof detection. To explore this rich information that is not considered yet in the literature, we present a novel tint rotation recognition task which focuses on the normal class colorimetry. Given an RGB image $\inImg$ and a transformation $\gamma$ where $\gamma(\inImg,\theta)$ adds an offset $\theta$ to the hue channel (in HSV space) of $\inImg$; we try to predict the distribution of $\Theta$ from $\gamma(\inImg,\Theta)$. For practical reasons, we limit the possible tint rotation angles to $c$ distributed angles and our task becomes to distinguish angles which are multiples of $\frac{2\pi}{c}$. 

Tackling the colorimetry task with a rotation detection task allows us to discriminatively learn high-scale and general colorimetry clues while keeping a low computational cost. In addition, we note that contrary to the geometrical rotation recognition task where a number of angles different from four would leave visual artifacts, our task does not have any limitation on $c$.

\begin{figure}[htb]
	\centering
	\begin{minipage}[b]{0.24\linewidth}
		\centering
		\centerline{\includegraphics[width=\linewidth]{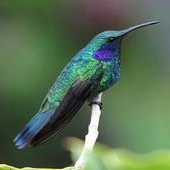}}
		\vspace{.2cm}
		\centerline{$\theta=0$°}
	\end{minipage}
	\begin{minipage}[b]{0.24\linewidth}
		\centering
		\centerline{\includegraphics[width=\linewidth]{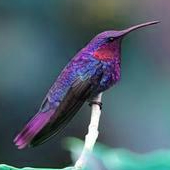}}
		\vspace{.2cm}
		\centerline{$\theta=90$°}
	\end{minipage}
	\begin{minipage}[b]{0.24\linewidth}
		\centerline{\includegraphics[width=\linewidth]{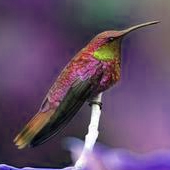}}
		\vspace{.2cm}
		\centerline{$\theta=180$°}
	\end{minipage}
	\begin{minipage}[b]{0.24\linewidth}
		\centerline{\includegraphics[width=\linewidth]{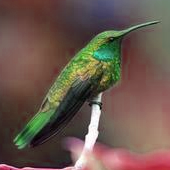}}
		\vspace{.2cm}
		\centerline{$\theta=270$°}
	\end{minipage}
	\caption{Tint rotation task for $c=4$ (\lblsec \ref{subsec:tint-rot}).}
	\label{fig:res}
\end{figure}

Nevertheless it is impossible to detect any tint rotation inside areas without any original color information. To prevent high anomaly scores on desaturated images, we need to give a lower weight on those regions. To this end, instead of working on the angle distribution we use the expected $L_1$ error in RGB space between the original image and the predicted one. Since we are computing a pixel wise RGB error, only large areas of colorful pixels will impact the anomaly score. The tint rotation task training loss is:
\begin{equation}
    \label{eqn:tint-loss}
	\lossTint(\gamma(\inImg,\theta))=\expect[\Theta|\gamma(\inImg,\theta)]{\frac{\|\inImg-\gamma(\inImg,\theta-\Theta)\|_1}{W\times H\times 255}}
\end{equation} where $W\times H$ is the dimension of the image. As for the anomaly score, we use the same error as the loss function which becomes in its developed form:
\begin{equation}
    \label{eqn:tint-score}
    \resizebox{1 \columnwidth}{!} {%
	\ensuremath{s_a(\gamma(\inImg,\theta))=\sum\limits_{i=1}^c \softmax{(\phi\circ f(\gamma(\inImg, \theta)))}_i \left(\frac{\|\inImg-\gamma(\inImg,\theta-i\cdot\frac{2\pi}{c})\|_1}{W\times H\times 255}\right)}
	}
\end{equation} where $\softmax(\cdot)$ is the softmax function.

By introducing this task we force our encoder to fully represent the normal class colorimetry, which could be potentially ignored by the puzzle task in case of salient geometrical features.

\subsection{Intra-piece tasks}
\label{subsec:intra-piece-tasks}

\begin{figure}[tbh]
	\centering
	\includegraphics[width=1\linewidth]{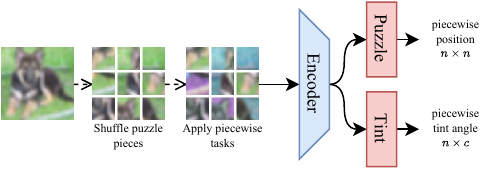}
	\caption{Example of intra-piece tasks with tint rotation detection with $c$ possible rotations (\lblsec \ref{subsec:intra-piece-tasks}). The only additional cost of this task when compared to the piece-wise task is a specialized dense layer.}
	\label{fig:intrapiece-tasks}
\end{figure}

On top of the piece-wise puzzle task, we further propose to add pretext sub-tasks inside each puzzle piece. Given an intra-piece task $\mathcal{T}_{piece}$ and an image composed of $n$ pieces images $\inImgPiece{1},\cdots,\inImgPiece{n}$, we first sample a random augmented piece using the pretext data generation function on each piece $(I^{(aug)}_i,y_i)\sim \datagen{\mathcal{T}_{piece}}(\{\inImgPiece{i}\})$. Then our network tries to solve simultaneously the puzzle task and the intra-piece tasks by minimizing the loss 
\begin{equation}
	\loss(\inImg)=\frac{1}{n}\sum_i^n \left(\crossentropy(\phi\circ f_i(\inImg); \Pi_i) + \lossIp(\inImgPiece{i})\right)
\end{equation} where the first term is from the piece-wise puzzle loss defined in \lbleq \ref{eqn:puzzle-loss} and $\lossIp$ is the loss of the intra-piece task. In our case, we choose the tint rotation task for the intra-piece task thus $\lossIp=\lossTint$. We argue that the piece-wise tint rotation task is more suitable than a piece-wise geometrical rotation task since it mixes different modalities rather than only combining geometrical cues. Besides, We have already studied the combination of jigsaw puzzle task with the geometric rotation in our previous work \cite{Jezequel2021}. A summary of the intra-piece task model is given in \lblfig \ref{fig:intrapiece-tasks}.

By adding these intra-piece tasks, we essentially consider $n$ new tasks during inference without increasing the number of forward pass in our encoder. The only cost is the additional specialized dense layer for the pretext task. Each intra-piece task will allow our network to focus on specific image patches.

One issue with this method is that we can potentially mix object pieces and background pieces. Solving tasks on background pieces would enable the model to generalize on image distribution far from the normal class object. As a result, we \textit{introduce a weight map} for each piece learned during training where higher weights are given for pieces covering the object. We could see this map as a rough segmentation of the normal object in the image. These are computed in a similar fashion as visual attention mechanism, which have previously successfully been used for learning weight maps for each pixels \cite{Wang2018}.

\begin{figure}[!tbh]
	\centering
	\includegraphics[trim=5pt 0 10pt 0, width=1\linewidth]{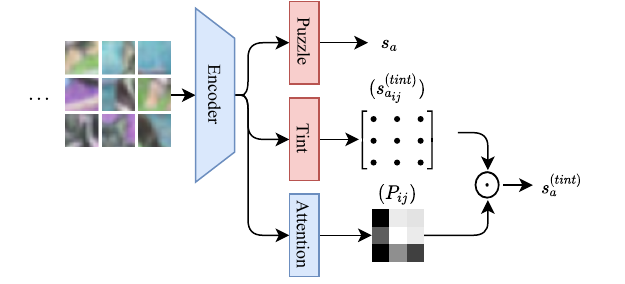}
	\caption{Intra-piece tasks with attention (\lblsec \ref{subsec:intra-piece-tasks}).}
	\label{fig:intrapiece-tasks-attention}
\end{figure}

First, we compute from the encoder representation $z$ a weight map $(w_{ij})_{0\leq i+j\leq n}$, which we normalize into attention weights using the $L_1$ normalized sigmoid $P_{ij}=\frac{\sigma(w_{ij})}{\|w\|_1}$.
This normalization function produces smoother maps than the classical softmax activation, preventing very sparse maps where only one piece has a non-null activation. To further prevent these cases, we include an additional term to the loss encouraging spread matrices:
\begin{align}
	\loss_{\text{density}}(P)&=\sum_{ij}\left\Vert \scriptsize{\begin{pmatrix}i\\j\end{pmatrix}}-\mu\right\Vert _{2}P_{ij}
\end{align} where $\mu=\sum_{ij}P_{ij}\scriptsize{\begin{pmatrix}i\\j\end{pmatrix}}$.

\textbf{Our final loss of intra-piece tasks} taking into account the attention map (upper branch in \lblfig\ref{fig:method-overview}) is:
\begin{equation}
	\lossupbranch=\lossPzl(\inImg)+\loss_{\text{density}}(P) + \sum_{i,j} P_{ij}\cdot \lossIp(\inImgPiece{i,j})
\end{equation} and the corresponding anomaly score is
\begin{equation}
	s_a(\inImg)=M(\left\{s_{OOD}((z,y);\phi\circ f)\middle|(z,y)\in \datagen{\mathcal{T}}(\{\inImg\})\right\})
\end{equation} where $M$ is the fusion function which is detailed in \lblsec\ref{sec:ood-methods}.

We can see in \lbltab \ref{tab:comp-attention} that the attention mechanism increases anomaly detection performances.

\subsection{Partial colorization task}
\label{subsec:part-color}

We present in this section a novel generative pretext task for anomaly detection which is highly texture oriented. In the colorization task commonly used in the literature \cite{Iizuka2016,Zhang2016ColorfulIC,Su2020}, the main objective is to predict the $(A,B)$ color channels from the luminance channel $L$ of an image in LAB space.

One big challenge with this task is to colorize the background since it can vary a lot inside the training normal set. The re-colorization will be naturally poorer for unseen background during inference of new observations. Therefore the object itself should have more impact on the AD algorithms than the background, making the anomaly detector more object-oriented than scene-oriented. In addition, several issues arise when considering the typical framework of colorization through regression \cite{Iizuka2016} where $\expect{(A_{ij}, B_{ij})|L}$ is directly estimated for each pixel $(i,j)$. First, the colorimetry of the normal class can potentially be multi-modal. In other words, the normal class objects can have several plausible set of colors called modes. For example, horses could have more than one fur color yet still being part of the same class. In this case a regression network will end up predicting the mean of all modes ignoring the multi-modality. Second, even if one of the object mode is correctly predicted, any error function will yield high values if the mode of the current observation is different.

To tackle these limitations, we establish a novel method to learn colorization well-suited to anomaly detection. First, we augment the available inputs with the color values of the image inside a border of size $\bsize$ to make the background re-colorization easier. For a simple unified background, our model will be encouraged to color areas near the center object similarly to the border areas and mitigate the background influence on AD. Our partial colorization task thus consists in predicting $(A,B)$ from the image with partial color channels $\Ipart=(L,A\odot M_{\bsize}, B\odot M_{\bsize})$ where $M_{\bsize}$ is a binary mask consisting of 1 in the border of size $\bsize$ and 0 in the center. Moreover, different to existing regression methods, we estimate the posterior density $p(A_{ij},B_{ij}|\Ipart)$ of each pixel to cover any color multi-modality. For density estimation, we explore two different ideas: (1) quantize the colors into a low-range discrete variable and perform multi-class classification; (2) parameterize the density with a gaussian mixture model and perform maximum likelihood estimation.

\subsubsection{Color bin classification}

By quantizing each color value into $K$ bins and assuming the two colors planes to be independent, we can define the resulting categorical variables by $2K$ probabilities: $P(A_{ij}=1),\cdots, P(A_{ij}=K), P(B_{ij}=1),\cdots, P(B_{ij}=K)$. We thus estimate a map $y$ of dimension $H\times W\times 2K$, where 
\begin{align}
	y_{i,j,2k}&=P(A_{ij}=k|\Ipart) \nonumber \\
	y_{i,j,2k+1}&=P(B_{ij}=k|\Ipart)
\end{align}
Inspired by the label smoothing idea \cite{Mueller2020}, a gaussian smoothing is applied to the output distributions in order to propagate our model confidence to neighbor color bins. Indeed we do not want to entirely penalize close color bins. As such the final estimated density $\hat{P}(A_{ij}|\Ipart)$ for a network $\phi$ is
\begin{equation}
	\hat{P}(A_{ij}=k|\Ipart)=\left(\softmax(\phi(\Ipart)_{ij})\star G_\sigma\right)_k
\end{equation} where $G_\sigma$ is the gaussian kernel of standard deviation $\sigma$.

\subsubsection{Gaussian Mixture Model MLE}

Our second approach is to parameterize the densities with Gaussian Mixture Models. Accordingly, we have for each pixel a sum of $K$ gaussian densities:
\begin{equation}
	p(A_{ij},B_{ij}|\Ipart)=\sum_{k=1}^K \pixclstr{\pi}{ij}{k} \normdist{A_{ij},B_{ij}}{\pixclstr{\mu}{ij}{k}}{\pixclstr{\Sigma}{ij}{k}}
\end{equation} where $\pixclstr{\pi}{ij}{k}\in\R$ is the prior probability of the $k$\textsuperscript{th} cluster, $\pixclstr{\mu}{ij}{k}\in\R^2$ is the mean color of the k\textsuperscript{th} cluster and $\pixclstr{\Sigma}{ij}{k}\in\R^{2\times 2}$ is the covariance color matrix of the $k$\textsuperscript{th} cluster.

Rather than predicting the full $2\times 2$ matrix $\pixclstr{\Sigma}{ij}{k}$, we only predict the three free parameters $\bm\sigma$. We can then reconstruct the positive definite covariance matrix using Cholesky decomposition \cite{Higham2009}:
\def\lmat{\left(\begin{matrix}1 & 0 \\ l & 1\end{matrix}\right)}
\begin{equation}
	\pixclstr{\Sigma}{ij}{k}=\lmat\text{Diag}\left(e^d\right)\lmat^T
\end{equation} where $d\in\R^2$ and $l\in\R$. This decomposition ensures strictly positive eigen values from the exponential and a semi-positive matrix from the Cholesky decomposition. All the possible covariance matrices are thus parameterized by $(d, l)$. It also introduces better numerical stability for determinant computation with the simple formula $\log|\Sigma|=\log\left|\text{Diag}\left(e^d\right)\right|=\sum_i d_i$.

To train this model, we could use as the loss function the log-likelihood which considers all pixels independent:
\begin{equation}
	\loss(\mu,\Sigma|A,B)=\sum_{ij}\log\left(\sum_{k=1}^{K}\pixclstr{\pi}{ij}{k}\normdist{A_{ij},B_{ij}}{\pixclstr{\mu}{ij}{k}}{\pixclstr{\Sigma}{ij}{k}}\right)
\end{equation}
However this function turns out to be very hard to directly optimize for each pixel and does not lead to any meaningful colorization. We use instead the classical Expectation Maximization algorithm. As for details, we carry out the three following steps:
\begin{enumerate}[label=({\small\textsc{step \arabic*}}), leftmargin=37pt]
	\item \textbf{Compute Mahalanobis distances}: \begin{equation}
	    \label{eqn:em-step1}
		\pixclstr{\Delta}{ij}{k}=\left(I_{ij}-\pixclstr{\mu}{ij}{k}\right)^{T}\pixclstr{\Sigma}{ij}{k}^{-1}\left(I_{ij}-\pixclstr{\mu}{ij}{k}\right)
	\end{equation}
	\item \textbf{Compute posterior cluster probabilities}: \begin{equation}
	    \label{eqn:em-step2}
		\gamma_{ij}(k)=\frac{\pixclstr{\pi}{ij}{k}\exp\left(-\frac{1}{2}\left(\sum_l\pixclstr{d}{l}{k} +  \pixclstr{\Delta}{ij}{k}\right)\right)}{\sum_{\kappa=1}^{K}\pixclstr{\pi}{ij}{\kappa}\exp\left(-\frac{1}{2}\left(\sum_l\pixclstr{d}{l}{\kappa} +  \pixclstr{\Delta}{ij}{\kappa}\right)\right)}
	\end{equation}
	\item \textbf{Fix the $\gamma_{ij}(k)$ and minimize loss} (lower branch): \begin{equation}
    	\label{eqn:color-loss}\hspace{-35pt}\resizebox{0.99 \columnwidth}{!} {\ensuremath{%
    	\losslowbranch(\pi,\mu,\Sigma|I)=\sum\limits_{ij}\sum\limits_{k=1}^{K}\gamma_{ij}(k)\left(\pixclstr{\Delta}{ij}{k}+\sum\limits_l \pixclstr{d}{l}{k}-\log\pixclstr{\pi}{ij}{k}\right)
    	}}
	\end{equation}
\end{enumerate}

Once the training is finished, we compute the anomaly score as the likelihood of the color channels under the predicted $\pixclstr{\pi}{ij}{k}$, $\pixclstr{\mu}{ij}{k}$ and $\pixclstr{\Sigma}{ij}{k}$:
\begin{equation}
    \label{eqn:color-score}
    s_a(A_{ij},B_{ij}|\Ipart)=\sum_{k=1}^K \pixclstr{\pi}{ij}{k} \normdist{A_{ij},B_{ij}}{\pixclstr{\mu}{ij}{k}}{\pixclstr{\Sigma}{ij}{k}}
\end{equation}

In order to choose the number of gaussians $K$, we apply beforehand a K-means color clusterization \cite{Jin2010} on the cropped down-sampled images of the normal class. Then by using the elbow method, we can find the optimal $K$ inside $\llbracket 1, 10 \rrbracket$.

\textbf{Advantages}. The GMM approach has three advantages over the bin classification: \textbf{(i)} its density support is not bounded, and is continuous thus not needing any gaussian smoothing, \textbf{(ii)} it can fully model the dependence between the color channels with the full covariance matrix, and \textbf{(iii)} it can reach the same quality of colorization with fewer parameters. The quality of colorization is here measured using the mean pixel color likelihood.

\begin{figure}[tbh]
	\centering
	\includegraphics[width=0.8\linewidth]{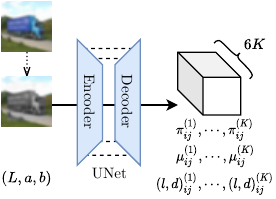}
	\caption{Scheme of the partial colorization with GMM estimation and a UNet network (\lblsec \ref{subsec:part-color}). The model predicts $6K$ parameters per pixel: $\pi\in\R$, $\mu\in\R^2$ and $\boldsymbol{\sigma}\in\R^3$ for each of the $K$ clusters.}
	\label{fig:pw-color-gmm}
\end{figure}

\begin{algorithm}[tbh]
	\caption{Our model training}\label{alg:training}
	
	\begin{algorithmic}[1]
		\STATE {\bfseries Input:} batch size $B$
		\STATE {\bfseries Initialization:} upper-branch encoder $\phi$, task-specific networks $f_{\text{pzl}}, f_{\text{tint}}$, attention network $f_{\text{att}}$, U-shape enc-dec $\psi$
		\WHILE{not reach the maximum epoch} 
		\STATE Sample image minibatch  $\x$
		\STATE Transform batch to $\nrs$ shuffled images $\x'_1,\cdots,\x'_{\nrs}$ with piecewise tint rotation
		\FOR{$k=1\cdots \nrs$}
			\STATE Apply encoder $\z_k\leftarrow\phi(\x'_k)$
			\STATE Compute puzzle loss $\lossPzl$ from \lbleqabbr \ref{eqn:puzzle-loss}
			\STATE Compute tint loss $\lossTint$ from \lbleqabbr \ref{eqn:tint-loss} with attention
		\ENDFOR
		\STATE Decolorize batch to $\x_\text{decolor}$
		\STATE Perform EM algorithm from \lbleqabbr \ref{eqn:em-step1},\ref{eqn:em-step2},\ref{eqn:color-loss}
		\STATE Gradient descent on $\lossupbranch$ to update $\phi,f_{\text{pzl}}, f_{\text{tint}}$, $f_{\text{att}}$
		\STATE Gradient descent on $\losslowbranch$ to update $\psi$
		\ENDWHILE
		\STATE {\bfseries Output:} {\bfseries networks $\phi, \psi, f_{\text{pzl}}, f_{\text{tint}}, f_{\text{att}}$}
	\end{algorithmic}
	
\end{algorithm}

\section{OOD methods and fusion}
\label{sec:ood-methods}

We try two different out-of-distribution methods for each pretext task: the softmax truth and the Mahalanobis distance. In the case of a self-supervised classification task, the most commonly used OOD function is the likelihood of the label given that the image is normal, which we call the ``\emph{softmax truth}'':
\begin{align}
	s_{OOD}((z, y);\phi\circ f)&=p(y|z,z\in\mathcal{X}_{train})\nonumber \\
	&\approx \softmax(\phi\circ f(z))_y
\end{align}

However, this softmax truth criterion takes into account only one component of the softmax vector. For easy tasks, we usually have a high probability on the correct class, however for harder, multi-issue task, we can have several typical highly activated classes for the normal class. As such, another idea is to look at the likelihood of the raw score vector given its label and given that the image is normal:
\begin{equation}
	s_{OOD}((z, y);\phi\circ f)=p(\phi\circ f(z)|y, z\in\mathcal{X}_{\text{train}})
\end{equation}
To approximate this conditional probability, the training dataset is first partitioned on samples sharing the same label value $l$, i.e. $\{(z,y)|(z,y)\in\mathcal{X}_{\text{train}}\text{ and } y=l\}$. The distribution of the normal class raw score vectors given $y$ can then be separately estimated on each partition after convergence of the network weights.

For a given classification problem with $C$ classes and a training set $\normset$, we estimate the mean scores $\mu_c$ and covariance matrices $\Sigma_c$ for each class $c$:
\begin{align}
	\mu_c&=\frac{1}{|\mathcal{Z}_c|}\sum_{z\in\mathcal{Z}_c} \phi\circ f(z) \nonumber \\
	\Sigma_c&=\frac{1}{|\mathcal{Z}_c|}\sum_{z\in\mathcal{Z}_c} (\phi\circ f(z)-\mu_c)^2
\end{align} where $\mathcal{Z}_c=\{z|(z,y)\in \datagen{\mathcal{T}}(\normset)\text{ and } y=c\}$.
The OOD score is approximated by the Mahalanobis distance \cite{McLachlan1999} with the mode corresponding to the truth label:
\begin{equation}
	s_{OOD}((z, y);\phi\circ f)\approx (\phi\circ f(z)-\mu_y)^T \Sigma_y^-1 (\phi\circ f(z)-\mu_y)
\end{equation}

We also explore different fusion functions to combine all the OOD scores into a single anomaly score. We first use the mean, but observe heavy biases from outlier OOD scores (very easy sub-task or harder sub-task). We then try different order statistics including the median and the 25\textsuperscript{th} percentile and compare the results in \lbltab \ref{tab:fusion-comp}.


\section{Full method overview}
\label{sec:method}
\begin{figure}[t]
	\centering
	\includegraphics[trim=0 0 15pt 0, width=1\linewidth]{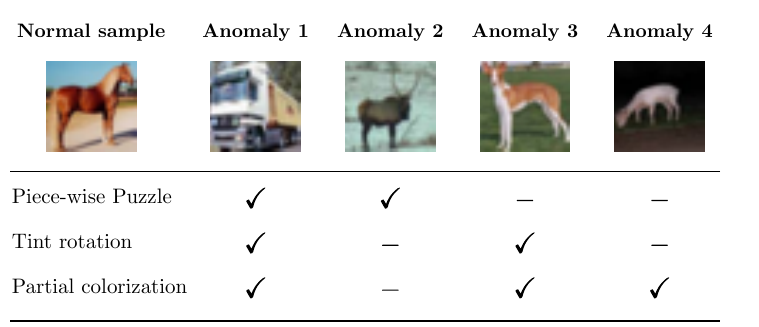}
	\caption{Examples of anomalies when considering one-vs-all on CIFAR-10. We indicate if each task detects it as an anomaly (\checkmark) or as normal ($-$).}
	\label{fig:anomaly-examples}
\end{figure}

This section summarizes our full method (\lblfig\ref{fig:method-overview}). Our model is made of two independent branches. The first discriminative branch (upper branch in \lblfig\ref{fig:method-overview}) solves the piece-wise puzzle task with intra-piece tint rotation detection task. The second generative branch (lower branch in \lblfig\ref{fig:method-overview}) performs the partial re-colorization task. We share the same encoder network for all of the discriminative tasks, including the attention mechanism. The re-colorization task is modeled with \emph{GMM}, and we include the \emph{attention mechanism} for the intra-piece task. To detect whether or not an observation $\x$ is an anomaly, we produce the OOD scores of the re-colorization and the $\nrstest$ sampled permutations along with tint rotation tasks. The chosen OOD function for every task is the \emph{softmax truth}. All of these scores are then combined into a single anomaly score using the \emph{median}. Our full training and inference algorithms are respectively given in \lblalg\ref{alg:training} and \lblalg\ref{alg:inference}.

\begin{algorithm}[tbh]
	\caption{Our model inference}\label{alg:inference}
	
	\begin{algorithmic}[1]
		\STATE {\bfseries Input:} image $\x$
		\STATE Transform input to $\nrstest$ shuffled images $\x'_1,\cdots,\x'_{\nrstest}$ with piecewise tint rotation
		\FOR{$k=1\cdots \nrstest$}
			\STATE Apply encoder $\z_k\leftarrow\phi(\x'_k)$
			\STATE Compute ${s_{\text{puzz}}}_k$ from \lbleqabbr \ref{eqn:puzzle-score}
			\STATE Compute ${s_{\text{tint}}}_k$ from \lbleqabbr \ref{eqn:tint-score}
		\ENDFOR
		\STATE Decolorize input to $\x_\text{decolor}$
		\STATE Compute $\boldsymbol{\mu}$,$\boldsymbol{\Sigma}$ and $\boldsymbol{\pi}$ using the U-shape enc.-dec. on $\x_\text{decolor}$
		\FOR{$i=1\cdots H, j=1\cdots W$}
		\STATE Compute ${s_\text{color}}_{i,j}$ from \lbleqabbr \ref{eqn:color-score}
		\ENDFOR
		
		\STATE \resizebox{0.96\columnwidth}{!}{$s_a\leftarrow \text{median}(\text{median}(s_{\text{puzz}_k}),\text{median}(s_{\text{tint}_k}),\text{median}(s_{\text{color}_{i,j}}))$}
		\STATE {\bfseries Output:} {\bfseries Anomaly score $s_a$}
	\end{algorithmic}
	
\end{algorithm}

Presented in \lblfig\ref{fig:anomaly-examples} are examples of anomalies detected by our three different tasks using different visual cues. As can be seen, our detectors are of complementary strength.

\section{Results}
\label{sec:results}

\subsection{Evaluation protocol}
\label{sec:protocols}

Our evaluation protocol is made of three types of anomaly detection challenges: object anomalies, fine-grained style anomalies, and face presentation attacks. First, to detect object anomalies we use general coarse object recognition datasets. The one-vs-all protocol is used, where we consider one class of a multi-classification dataset as the normal class. All the other classes are then considered as anomalous, and we can obtain a set of runs for each possible normal class. Thus, for a given run the training dataset is the normal class training data and the test dataset contains the original test data of the normal class and the anomalous classes. The final reported result is the mean of all runs.

However, these datasets have become far from real anomaly detection applications and might not be enough to fully evaluate AD methods. Thus we include a second evaluation group where we try to detect style anomalies using fine-grained classification datasets. Fine-grained datasets have been introduced to tackle the recognition of classes, usually part of a same category, with slight differences. We use here the one-vs-all protocol as well.

Finally, we consider a real anomaly detection problem which incorporates object anomalies, style anomalies and local anomalies. In particular we choose a dataset from face presentation attack detection (FPAD), where the goal is to discriminate real faces from fake representations of someone's face. Due to the constantly evolving frauds and high variability, anomaly detection seems a very appealing solution to this problem.

We use the following datasets:

\textbf{(i)} For \textbf{object anomalies}:
\begin{itemize}
	\item \textbf{F-MNIST} \cite{Xiao2017FashionMNISTAN}: has been introduced as a harder version of MNIST with 10 different classes of fashion items. All images are grayscale meaning no color information can be used to discriminate anomalies.
	\item \textbf{CIFAR-10} \cite{Krizhevsky2009LearningML}: object recognition dataset composed of 10 wide classes with 6000 images per class.
	\item \textbf{CIFAR-100} \cite{Krizhevsky2009LearningML}: extended version of CIFAR-10 with 100 classes each containing 600 images.
\end{itemize}

\textbf{(ii)} For \textbf{style anomalies}:
\begin{itemize}
	\item \textbf{Caltech-UCSD Birds 200} \cite{WelinderEtal2010}: fine-grained classification dataset of 200 birds species with approximately 30 images per class.
	\item \textbf{FounderType-200} \cite{10.1109/CVPR.2017.439}: font recognition dataset containing 200 fonts with 6700 images per class. It has been introduced for novelty detection and even though these images lie on a low dimensional manifold compared to natural images, they still provide insight into how well the model can capture small shape hints.
\end{itemize}

\begin{figure}[tbh]
	\centering
	\begin{minipage}[b]{0.19\linewidth}
		\centering
		\centerline{\includegraphics[width=\linewidth]{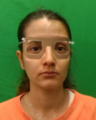}}
		\vspace{-3pt}\centerline{\captionfont (a)}
	\end{minipage}
	\begin{minipage}[b]{0.19\linewidth}
		\centering
		\centerline{\includegraphics[width=\linewidth]{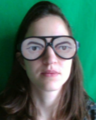}}
		\vspace{-3pt}\centerline{\captionfont (b)}
	\end{minipage}
	\begin{minipage}[b]{0.19\linewidth}
		\centering
		\centerline{\includegraphics[width=\linewidth]{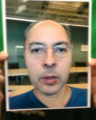}}
		\vspace{-3pt}\centerline{\captionfont (c)}
	\end{minipage}
	\begin{minipage}[b]{0.19\linewidth}
		\centering
		\centerline{\includegraphics[width=\linewidth]{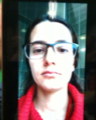}}
		\vspace{-3pt}\centerline{\captionfont (d)}
	\end{minipage}
	\begin{minipage}[b]{0.19\linewidth}
		\centering
		\centerline{\includegraphics[width=\linewidth]{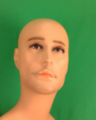}}
		\vspace{-3pt}\centerline{\captionfont (e)}
	\end{minipage}
	\begin{minipage}[b]{0.19\linewidth}
		\centering
		\centerline{\includegraphics[width=\linewidth]{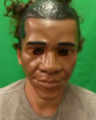}}
		\vspace{-3pt}\centerline{\captionfont (f)}
	\end{minipage}
	\begin{minipage}[b]{0.19\linewidth}
		\centering
		\centerline{\includegraphics[width=\linewidth]{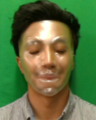}}
		\vspace{-3pt}\centerline{\captionfont (g)}
	\end{minipage}
	\begin{minipage}[b]{0.19\linewidth}
		\centering
		\centerline{\includegraphics[width=\linewidth]{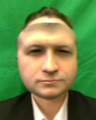}}
		\vspace{-3pt}\centerline{\captionfont (h)}
	\end{minipage}
	\begin{minipage}[b]{0.19\linewidth}
		\centering
		\centerline{\includegraphics[width=\linewidth]{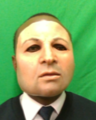}}
		\vspace{-3pt}\centerline{\captionfont (i)}
	\end{minipage}
	\begin{minipage}[b]{0.19\linewidth}
		\centering
		\centerline{\includegraphics[width=\linewidth]{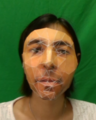}}
		\vspace{-3pt}\centerline{\captionfont (j)}
	\end{minipage}
	\caption{Overview of the WMCA dataset with 347 bonafide, \textbf{style anomalies} made of 200 print (c), 348 replay (d), 122 fake head (e), 137 rigid mask (f)(g)(h), 379 flexible mask (i), 71 paper mask (j) and \textbf{local anomalies} made of 75 face glasses (a)(b).}
	\label{fig:wmca-samples}
\end{figure}

\textbf{(iii)} For the \textbf{face presentation attack detection}, we use the \textbf{WMCA} dataset \cite{George2020} which contains more than 1900 short videos of real faces and presentation attacks. It contains several modalities such as infra-red or depth, but here we only use RGB. There are 72 real identities along with several types of attacks: paper print, screen replay, masks and partial attacks where only a localized area of the face is fake. The masks are composed of paper masks, rigid mask and flexible masks. An example of each type of attack is given in \lblfig \ref{fig:wmca-samples}.

\begin{table}[tbh]
	\caption{\label{tab:db-sum}Summary of evaluation datasets.}
	\centering
	\hspace{-1.5cm}
	\begin{tabular}{rl|ccc}
		\tablectop{2-5}
		&  \multirow{2}{*}{\textbf{Dataset}} & \multicolumn{3}{c}{\textbf{Anomaly type}} \\
		&   & Object     & Style      & Local      \\ \cline{2-5} 
		\groupcell{3}{1.7cm}{1.55cm}{\hfill \scriptsize Obj.classif}  & F-MNIST       & \checkmark &  \crossmark          & \crossmark  \\
		& CIFAR-10      & \checkmark &  \crossmark   & \crossmark  \\
		& CIFAR-100     & \checkmark &  \crossmark & \crossmark \\ \cline{2-5} 
		\groupcell{2}{1.7cm}{1.55cm}{\hfill \scriptsize Fine-grained} & Caltech-Birds & \checkmark & \checkmark & \crossmark \\
		& FounderType   & \crossmark & \checkmark &  \crossmark \\ \cline{2-5} 
		\groupcell{1}{1.77cm}{1.66cm}{\vspace{-5pt}\flushright \scriptsize FPAD} & WMCA          & \checkmark & \checkmark & \checkmark \\
		\tablecbottom{2-5}
	\end{tabular}
\end{table}

In all evaluations, the metric used is the area under the ROC curve (\textbf{AUROC}) or the error 1-AUROC, averaged over all possible normal classes in the case of one-vs-all datasets. We additionally include for anti-spoofing datasets metrics more adapted to biometric presentation attack detection:
\begin{itemize}
	\item The equal error rate (\textbf{EER}\cite{Chingovska2019}), which is the location in the ROC curve where the false reject rate (or Bona-fide Presentation Classification Error Rate BPCER) is equal to the false acceptance rate (or Attack Presentation Classification Error Rate APCER).
	\item The Attack Presentation Classification Error Rate for the Bona-fide Presentation Classification Error Rate fixed at 5\% (\textbf{APCER@5\%BPCER} \cite{Chingovska2019}).
\end{itemize}

\subsection{Implementation details}
\label{subsec:implementation}

\input{sota1.tex}
\input{sota2.tex}

For the piece-wise puzzle task, we use a margin of half the size of the pieces and find best results with $\nrstest=18$. Generally we use $n_w=n_h=3$ pieces for most datasets, except face anti-spoofing where $n_w=3$ and $n_h=4$. We observe better results with more vertical pieces on faces, since they are always upright and need finer vertical analysis. For the tint rotation recognition we use $c=4$ and for the re-colorization task, we use a contextual border $\bsize$ of two pixels.

Regarding network architecture, we use a 16-4 WideResNet \cite{Zagoruyko2016WideRN} ($\approx10M$ parameters with a depth of 16) for the feature extractor network $\phi$, along with three dense layers respectively of size $n^2$ for the piece-wise puzzle task, size $n\cdot c$ for the tint rotation task and size $n$ for the attention. Each of these dense layers have a dropout rate of 0.3 \cite{Srivastava2014DropoutAS}. As for the re-colorization task, we use a UNet network \cite{Ronneberger2015}. It was originally introduced for image segmentation, using a down-sample / up-sample strategy reintroducing the intermediate maps at each step of the down-sample branch into the up-sample branch. It is in fact generally well suited for any prediction task where the output is aligned with the input pixels (in our case a vector of GMM parameters for each pixel). Training is performed under SGD optimizer with Nesterov momentum \cite{Sutskever2013OnTI}, using a batch size of 32 and a cosine annealing learning rate scheduler \cite{Loshchilov2017}.
\vspace{-0.035cm}

\subsection{Comparison to the state-of-the-art}
\label{subsec:comp-sota}

A comparison of our method with other state-of-the-art (SOTA) anomaly detection models is performed on all three protocols. We choose to include three families of SOTA methods: \emph{one-class learning} methods which only learn using the normal class, \emph{semi-supervised learning} methods where a small set of anomalies is used during training and \emph{supervised learning}. The considered one-class methods can be categorized into \textbf{(1)} reconstruction error-based methods with ADGAN \cite{A:deecke2018image}, GANomaly \cite{akcay2018ganomaly} and PIAD \cite{Tuluptceva2019PerceptualIA}, \textbf{(2)} hybrid methods with OCSVM \cite{Schlkopf1999SupportVM}, IF \cite{isolationforest2009} , OC-CNN \cite{Oza2019}, \textbf{(3)} pretext tasks-based methods with ARNet \cite{Fei2020}, GeoTrans \cite{Golan2018DeepAD}, MHRot \cite{Hendrycks2019UsingSL} and PuzzleGeom \cite{Jezequel2021} and \textbf{(4)} two-stage anomaly detection using contrastive learning with SSD \cite{sehwag2021ssd} and DROC-contrastive \cite{DBLP:conf/iclr/SohnLYJP21}. GeoTrans uses various geometrical transformations as SSL pretext task, MHRot adds on top 90° rotations and our previous model PuzzleGeom \cite{Jezequel2021} includes a basic jigsaw puzzle task. 
Regarding semi-supervised methods, we evaluate DeepSAD \cite{deepsad2020} trained on the same normal samples but with three different ratio of the anomaly sub-classes: 10\%, 25\% and 75\%. For the fully supervised baseline we simply use the same backbone as our one-class method (the 16-4 WideResNet) extended with a dense layer representing the two normal and anomaly classes. It is important to note that its training is performed with classical binary cross-entropy loss on the normal/anomaly label, without any class balancing mechanism.

The experiment results are displayed in \lbltab \ref{tab:sota-table} and a detailed evaluation on the CIFAR-10 dataset is included in \lbltab \ref{tab:sota-table3}. We note that for the sake of fair comparison in the same conditions, we re-evaluate almost all methods ourselves using existing implementations. 

Our method maintains among the best accuracies on coarse object and fine-grained anomaly detection. It improves upon PuzzleGeom, and closes the gap toward semi-supervised performances with a small AUC difference of 0.5\% on CIFAR-100. Compared to previous pretext tasks such as rotation detection, our proposed tasks can better focus on local parts of the image. The re-colorization task will target more fine-grained local textures while the puzzle task and intra-piece tint detection will work on higher-scale geometrical and colorimetric features of the image. We also show that our method greatly improves anti-spoofing detection performance on WMCA. It even outperforms the supervised model and semi-supervised anomaly detection methods which have access up to 75\% of the anomalous data.

In general we can notice that hybrid methods, although efficient for smaller problems, do not extend well to high-dimensional data. The evaluated reconstruction-based methods also tend to fall behind pretext-task oriented models. On the other hand, two-stage contrastive methods like DROC-contrastive produce very competitive performance. This model combines different techniques including contrastive representation learning, distribution augmentation and OC-SVM. It performs slightly better than ours on the F-MNIST dataset and reaches the same AUC on CIFAR-10 but on the more challenging one, CIFAR-100, we obtain a gain of nearly 2\%. Moreover, we note that distribution augmentation and OC-SVM could also be used on the concatenation of our learned representations to reach better accuracy.

Overall, our model keeps a good balance between coarse object anomaly detection and finer style anomaly detection, and even outperforms semi-supervised anomaly detection methods on CUB-200 and WMCA. It achieves a relative error improvement of 36\% on CIFAR-10 and 40\% on WMCA compared to PuzzleGeom.

\begin{table}[!htbp]
	\caption{\label{tab:sota-table2}AUROC, EER and APCER at 5\% BPCER on WMCA dataset, best result is in bold.}
	\centering
	\begin{tabular}{l|ccc}
		\tabletop
		\textbf{Models}     & \textbf{AUROC} & \textbf{EER} & \textbf{\begin{tabular}[c]{@{}c@{}}APCER\\ (5\%BPCER)\end{tabular}}  \\ 
		\hline
		MHRot \cite{Hendrycks2019UsingSL} & 81.3           & 23.9                & 72.6\%            \\
		PuzzleGeom \cite{Jezequel2021} & 85.6           & 19.7                & 33.8\%            \\
		\hdashline
		Ours                & \textbf{91.4} & \textbf{16.1}  & \textbf{27.3\%}  \\
		\tablebottom
	\end{tabular}
\end{table}

Lastly, we compare in \lbltab \ref{tab:sota-table2} our method with the two second best self-supervised methods MHRot and PuzzleGeom on WMCA. Using our method the APCER@5\%BPCER drops from 33.8\% to 27.3\%. This also shows promising usage of anomaly detection methods in fraud detection.

\section{Parameter study}
\label{sec:param-study}

In this section, we evaluate the parametrization of pretext tasks in \lblsecs\ref{subsec:param_puzzle}, \ref{subsec:param_tint}, \ref{subsec:param_color}, the choice of OOD function in \lblsec \ref{subsec:param_ood} and perform an ablation study in \lblsec \ref{subsec:param_abl_att}.

\subsection{Puzzle task complexity}
\label{subsec:param_puzzle}

\begin{figure}[tbh]
	\centering
	\resizebox{1.06\linewidth}{!}{
		\hspace{-25pt}\input{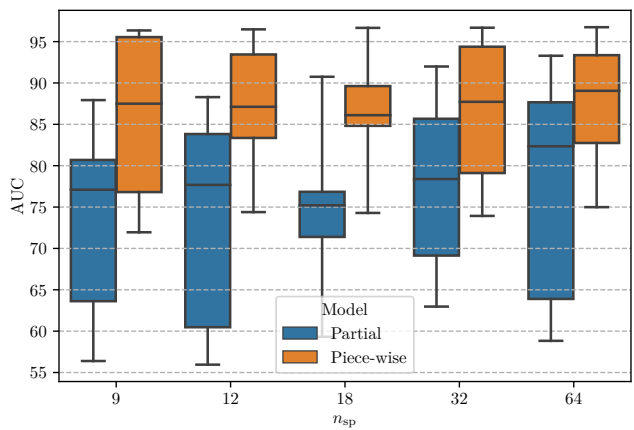}
	}
	\vspace{-20pt}
	\caption{\label{fig:puzzle-comp} Comparison of AUC with different number of tested permutations $\nrstest$ for 3x3 partial and piece-wise puzzle on CIFAR-10 dataset.}
\end{figure}

We start by comparing in \lblfig \ref{fig:puzzle-comp} the two approaches on the CIFAR-10 dataset for the jigsaw puzzle task introduced in \lblsec \ref{subsec:pw-puzzle}. The piece-wise puzzle task greatly improves performances for all CIFAR-10 classes even though the same permutations are tested during inference. Moreover, we confirm that the partial puzzle task is more sensitive to the choice of $\nrstest$, since its representation quality also depends on this factor. We choose to fix $\nrstest=18$ out of $9!$ possible permutations as a good compromise between complexity of inference and accuracy. 

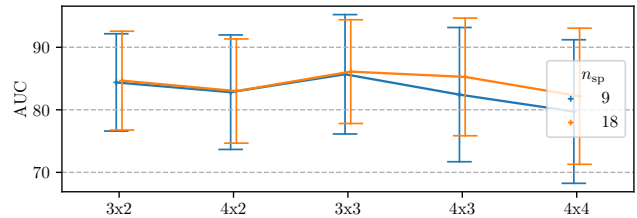
\begin{figure}[tbh]
	\centering
	\resizebox{1.06\linewidth}{!}{
		\hspace{-25pt}\input{figures/puzzle_n_piece_comp.pgf}
	}
	\vspace{-20pt}
	\caption{\label{fig:puzzle-n-piece-comp} Comparison of the number of pieces on CIFAR-10 dataset with two different amounts of permutations during inference.}
\end{figure}

The influence of the number of puzzle pieces $n_w$ and $n_h$ for $\nrstest\in\{9,18\}$ is reported in \lblfig \ref{fig:puzzle-n-piece-comp} on CIFAR-10. We can see that for both $\nrstest=9$ and $\nrstest=18$, the best value for general one-vs-all problem is $n_w=n_h=3$.

\subsection{Tint rotation task complexity}
\label{subsec:param_tint}

\begin{figure}[tbh]
	\centering
	\resizebox{1.06\linewidth}{!}{
		\hspace{-25pt}\input{figures/hue_rot_comp_auc_plot.pgf}
	}
	\vspace{-20pt}
	\caption{\label{fig:comp-tint-rot}Comparison of AUC with different number of tint rotation $c$ on CIFAR-10 dataset.}
\end{figure}
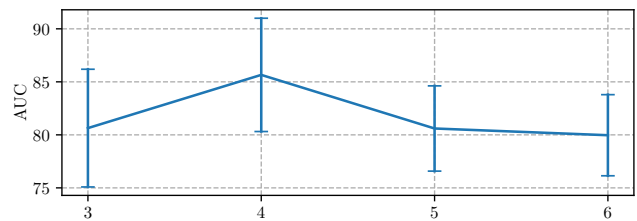

We measure the AUC of the isolated tint rotation task for different number of tint rotations $c$ on the CIFAR-10 dataset in \lblfig\ref{fig:comp-tint-rot}. The best value of $c$ across several normal classes is 4.

\subsection{Colorization task parametrization}
\label{subsec:param_color}

The two colorization parametrizations using Gaussian Mixture Model and bin classification are compared on the normal class full colorization task. Our evaluation metric is directly the likelihood of the colorization, which is respectively for classification and GMM
\begin{equation}
	\loss(A,B)=\prod_{i,j}\softmax(\phi(I)_{ij})_{\floor{\frac{A_{ij}}{K}}}\cdot\softmax(\phi(I)_{ij})_{\floor{\frac{B_{ij}}{K}}}
\end{equation} and 
\begin{equation}
	\loss(A,B)=\prod_{i,j}\sum_{k=1}^{K}\pixclstr{\pi}{ij}{k}\normdist{A_{ij},B_{ij}}{\pixclstr{\mu}{ij}{k}}{\pixclstr{\Sigma}{ij}{k}}
\end{equation}

\begin{figure}[tbh]
	\centering
	\begin{minipage}[b]{0.47\linewidth}
		\vspace{.2cm}
		\centering
		\centerline{\includegraphics[width=0.5\linewidth]{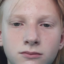}}
		\centerline{\captionfont original}
		\vspace{.2cm}
		\begin{minipage}[b]{0.45\linewidth}
			\centerline{\includegraphics[width=\linewidth]{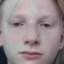}}
			\centerline{\captionfont bin class.}
		\end{minipage}
		\begin{minipage}[b]{0.45\linewidth}
			\centerline{\includegraphics[width=\linewidth]{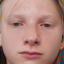}}
			\centerline{\captionfont GMM}
		\end{minipage}
		\vspace{.2cm}
	\end{minipage}%
	\hspace{0pt}
	\vrule
	\hspace{1.2pt}
	\begin{minipage}[b]{0.47\linewidth}
		\centering
		\centerline{\includegraphics[width=0.5\linewidth]{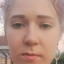}}
		\centerline{\captionfont original}
		\vspace{.2cm}
		\begin{minipage}[b]{0.45\linewidth}
			\centerline{\includegraphics[width=\linewidth]{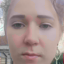}}
			\centerline{\captionfont bin class.}
		\end{minipage}
		\begin{minipage}[b]{0.45\linewidth}
			\centerline{\includegraphics[width=\linewidth]{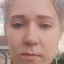}}
			\centerline{\captionfont GMM}
		\end{minipage}
		\vspace{.2cm}
	\end{minipage}
	\caption{Colorization comparison on faces. The first row displays the original images, while the second represents the re-colorization of two methods. As we can see, the bin classification approach produces dull colors and mixes the skin color modes, producing grayish colors. }
	\label{fig:colorization-comp}
\end{figure}

Overall, we can reach higher likelihoods with GMM than bin classification. Moreover, a better separation of the different modes can be achieved using GMM, where bin classification usually mixes the different modes and produces dull colors (see \lblfig \ref{fig:colorization-comp}).

\subsection{Choice of OOD and fusion functions}
\label{subsec:param_ood}

To evaluate the effect of Mahalanobis distance as an anomaly score, we compare it with the softmax truth and its improved form, the ODIN method \cite{Liang2018} which adds temperature scaling during training, and the input pre-processing $\tilde{\x}=\x-\varepsilon\operatorname{sign}(-\nabla_{\x}\log \softmax(\x;T))$.

\begin{figure}[tbh]
	\centering
	\resizebox{1.125\linewidth}{!}{
		\hspace{-28pt}\input{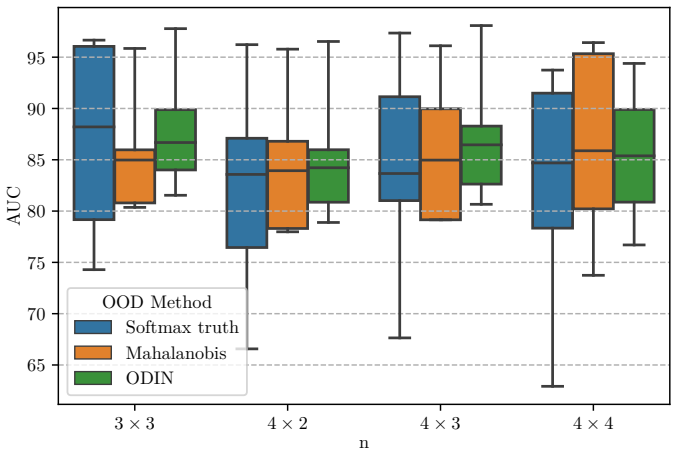}
	}
	\vspace{-20pt}
	\caption{\label{fig:ood-comp} Comparison of OOD methods AUC with different number of pieces $n$ for $\nrstest=18$ tested permutations on CIFAR-10 dataset.}
\end{figure}

The results are presented in \lblfig \ref{fig:ood-comp} for different number of puzzle pieces $n$ and $\nrstest=18$ permutations tested. The AUC increases with the number of pieces when using the Mahalanobis distance, whereas it decreases with the softmax truth. In addition, the AUC of the most difficult class is always higher when using the Mahalanobis distance. This shows that despite a lower average anomaly detection performance, it has less variance in its predictions and provides more robust OOD scores to different normal classes. Even though the ODIN method provides sensible improvement for more than $3\times 3$ pieces, it greatly increases computational complexity during training and inference. In our tests, we observe an inference time increase of more than three times with the ODIN method. We provide in \lbltab\ref{tab:comp-mahalanobis} further comparisons between the softmax truth and the Mahalanobis distance on the puzzle task with $\nrstest=9$.

\begin{table}[tbh]
	\caption{\label{tab:comp-mahalanobis}Comparison of AUC with different OOD methods for the piece-wise puzzle task with $\nrstest=9$ on CIFAR-10 dataset.}
	\centering
	\begin{tabular}{c|c|c|c|c}
		\tabletop
		\textbf{n} & \textbf{OOD Method} & \boldmath$\mu_{AUC}$ & \textbf{max\textsubscript{AUC}} & \textbf{min\textsubscript{AUC}} \\
		\hline 
		\multirow{2}{*}{$3\times 3$} & Softmax truth & 86.39 & 96.35 & 71.95 \\
		& Mahalanobis & 83.44 & 95.30 & 80.60 \\
		\hline 
		\multirow{2}{*}{$4\times 2$} & Softmax truth & 82.82 & 96.11 & 65.49 \\
		& Mahalanobis & 83.21 & 95.74 & 80.10 \\
		\hline 
		\multirow{2}{*}{$4\times 3$} & Softmax truth & 84.13 & 96.34 & 66.57 \\
		& Mahalanobis & 84.41 & 96.08 & 81.22 \\
		\hline 
		\multirow{2}{*}{$4\times 4$} & Softmax truth & 80.61 & 93.48 & 61.87 \\
		& Mahalanobis & 86.58 & 96.29 & 80.00 \\
		\tablebottom
	\end{tabular}
\end{table}

Finally, we evaluate the choice of different fusion functions on the WMCA dataset in \lbltab \ref{tab:fusion-comp}. The evaluated fusion functions are simple order statistics commonly found among ensemble learning decision fusion strategies. We observe overall better performances regarding AUC and APCER with the median fusion function.

\begin{table}[tbh]
	\caption{\label{tab:fusion-comp}Comparison of AUC and APCER@5\%BPCER with different fusion functions for the puzzle task on WMCA dataset.}
	\centering
	\begin{tabular}{c|c|c}
		\tabletop
		\textbf{Function} & \textbf{AUC} & \textbf{\begin{tabular}[c]{@{}c@{}}APCER\\ (5\%BPCER)\end{tabular}} \\
		\hline 
		Mean & $90.12\pm 0.42$ & 30.3\\
		25\textsuperscript{th} percentile & $91.63\pm 0.50$ & 29.2 \\
		Median & $91.41\pm 0.45$ & 27.3\\
		\tablebottom
	\end{tabular}
\end{table}

\subsection{Ablation study}
\label{subsec:param_abl_att}

We evaluate the impact of each pretext task on the final anomaly detection AUROC. In \lbltab\ref{tab:abl-study}, we compare on CIFAR-10 the basic partial puzzle model with the addition of the piece-wise puzzle task, colorization task, intra-piece tint rotation detection task with and without the attention map. While the piece-wise puzzle and colorization give our model great discrimination power with an AUC of 89.12, the intra-piece task with attention further refines our model.

\begin{table}[tbh]
	\caption{\label{tab:abl-study}Ablation study of each component on CIFAR-10 using the AUROC. The baseline is the partial puzzle task.}
	\centering
	\begin{tabular}{cccc|c}
		\tabletop 
		\multicolumn{4}{c|}{\textbf{Ablation Settings}}                                                                    & \multirow{2}{*}{\vspace{-10pt}\textbf{AUC}} \\
		\scriptsize \makecell[tc]{Piece-wise \\ puzzle} & \scriptsize Colorization & \scriptsize \makecell[tc]{Intra-piece \\ tint rotation} & \scriptsize Attention &                      \\ \hline
		\crossmark             & \crossmark               & \crossmark                   & \crossmark            & 75.44                \\
		\checkmark             & \crossmark               & \crossmark                   & \crossmark            & 86.97                \\
		\checkmark             & \checkmark               & \crossmark                   & \crossmark            & 89.12                \\
		\checkmark             & \checkmark               & \checkmark                   & \crossmark            & 90.94                \\
		\checkmark             & \checkmark               & \checkmark                   & \checkmark            & 92.48                \\
		\tablebottom
	\end{tabular}
\end{table}

We also investigate on more datasets how the addition of attention in the intra-piece task improves anomaly detection in \lbltab \ref{tab:comp-attention}. By including attention weights for each piece, we can further improve the mean AUC on all datasets, although marginally increasing the prediction variances on different normal classes. We can also notice that the usage of attention has varying contribution depending on the dataset. The main role of the attention for the intra-piece task is to prevent our task-specific model to generalize too much on background pieces. Thus, attention will benefit the most when the normal class background is very diverse or the normal object is very small in the image.

\begin{table}[tbh]
	\caption{\label{tab:comp-attention}Ablation study of the intra-piece task attention using the AUROC.}
	\centering
	\begin{tabular}{c|ccc}
		\tabletop 
		\multirow{2}{*}{\scriptsize \textbf{Att.}} & \multicolumn{3}{c}{\scriptsize \textbf{AUC}}                                      \\
		& CIFAR10         & CIFAR100        & WMCA            \\ \hline
		\crossmark                     & $90.94\pm 0.51$ & $88.06\pm 0.84$ & $90.29\pm 0.34$ \\
		\checkmark                     & $92.48\pm 0.52$ & $88.21\pm 0.83$ & $91.43\pm 0.35$ \\
		\tablebottom
	\end{tabular}
\end{table}

\section{Conclusion and Future Work}
\label{sec:conclusion}

We explore in this paper more efficient pretext tasks and show that a combination of a colorization and a puzzle task with intra-piece tint rotation subtasks provides the best anomaly detection performances. We also show the importance of different out-of-distribution functions along with their fusion functions. Finally, we provide a more comprehensive evaluation protocol than previously used datasets in the anomaly detection literature. It presents more challenging datasets and covers object, style and local anomalies. Our method outperforms state-of-the-art, including a semi-supervised method, on most of the fine-grained datasets.

For future work we could explore other generative pretext tasks such as image reconstruction. As in the colorization task, only a part of the image mostly covering the normal object would be destroyed. Furthermore, generative tasks such as our current colorization could be used to locate anomalies using the pixel-wise error. Finally we could reframe our method into a two-stage anomaly detection. In a first step, representations would be learned solving our pretext re-colorization, jigsaw puzzle and intra-piece tint rotation detection tasks. Then we could separately train a OC-SVM on the concatenation of representations from the puzzle and colorization encoder. We could further evaluate our model with differently sized backbones and measure the impact on each of our three pretext tasks.

\bibliographystyle{IEEEtran}
\bibliography{IEEEnolink, IEEEabrv, refs}

\end{document}

%% file: sota1.tex
\begin{table*}[t]
	\caption{\label{tab:sota-table}Comparison with the state-of-the-art AUROC over several datasets, underline indicates best result, bold indicates best one-class learning result. For the sake of fair comparison, we re-evaluated by ourselves all methods, except the one-class methods in the first block {\scriptsize (results are from \cite{A:deecke2018image,Fei2020,DBLP:conf/iclr/SohnLYJP21})}. DROC-contrastive \cite{DBLP:conf/iclr/SohnLYJP21} combines different techniques: contrastive learning, distribution augmentation and OC-SVM.}
	\centering
	\hspace{-1.2cm}
	\begin{tabular}{rl||ccc|cc|cc}
		\tablectop{2-8}\rule{0px}{10px}
		& \textbf{Model}           & \textbf{CIFAR-10} & \textbf{CIFAR-100}     & \textbf{F-MNIST}       & \textbf{CUB-200}               & \textbf{FounderType}                    & \textbf{WMCA}  \\ 
		\cline{2-8}
		\groupcell{1}{1.9cm}{1.7cm}{\hfill\tiny Supervised} & 16-4 WideResNet \cite{Zagoruyko2016WideRN} & 99.3        & 96.3          & 99.2  & -   & -  & 82.4                         \\
		\cline{2-8}
		\groupcell{3}{1.9cm}{1.7cm}{\flushright\tiny Semi-Supervised} & Deep-SAD (75\%) \cite{deepsad2020} & \underline{92.5}        & 88.7          & \underline{98.1}  & 73.6                  & \underline{99.8}                  & 83.2                         \\
		& Deep-SAD (25\%) &  90.8    & 87.9          & 95.4          & 70.9                  & 99.4                     & 79.8                         \\
		& Deep-SAD (10\%) & 86.0         & \underline{89.1}  & 88.2          & 66.1                  & 98.0                       & 72.6                         \\ 
		\cline{2-8}
		\groupcell{11}{1.9cm}{1.7cm}{\hfill\tiny One-class} & ADGAN \cite{A:deecke2018image} & 62.4 & 54.7 & 88.4 & - & - & - \\
		& GANomaly \cite{akcay2018ganomaly} & 69.5 & 56.5 & 80.9 & - & -& - \\
		& ARNet \cite{Fei2020} &  86.6 & 78.8 & 93.9 & - & - & - \\
		& DROC-contrastive \cite{DBLP:conf/iclr/SohnLYJP21} & \underline{\textbf{92.5}} & 86.5 & \textbf{94.8} & - & - & - \\
		\cdashline{2-8}
		
		& OCSVM \cite{Schlkopf1999SupportVM} & 58.5 & - & 74.2 & 76.3 & -            & -                  \\
		& IF \cite{isolationforest2009} & 73.4 & - & 84.0 & 74.2 & -& -\\
		& OC-CNN \cite{Oza2019}          & 66.5        & -          & 75.4 & -                  & -                   & -                         \\
		
		& PIAD \cite{Tuluptceva2019PerceptualIA} & 79.9        & 78.8          & 94.3 & 63.5                  & 90.8                      & 76.4                         \\
		& GeoTrans \cite{Golan2018DeepAD} & 85.4        & 84.7          & 92.6          & 66.6                  & 92.3                   & 79.8                         \\
		
		& MHRot \cite{Hendrycks2019UsingSL} & 89.5        & 83.6          & 92.5          & 77.6                  & 96.7                   & 81.3                         \\
		& PuzzleGeom \cite{Jezequel2021} & 88.2        & 85.8 & 92.8          & \underline{\textbf{83.2}} & 96.9 & 85.6                         \\ 
		\cdashline{2-8}
		& Ours            & \underline{\textbf{92.5}}        & \textbf{88.2}             & 93.7             & \underline{\textbf{83.2}}                     & \textbf{97.4}             &  \underline{\textbf{91.4}}                        \\
		\tablecbottom{2-8}
	\end{tabular}
\end{table*}

%% file: sota2.tex
\begin{table*}[!htbp]
	\caption{\label{tab:sota-table3}Detailed comparison with one-class state-of-the-art AUROC on the CIFAR-10 dataset.}
	\centering
	\begin{tabular}{l|cccccccccc|c}
		\tabletop
		\textbf{Model}    & \textbf{Airplane} & \textbf{Automobile} & \textbf{Bird} & \textbf{Cat}  & \textbf{Deer} & \textbf{Dog}  & \textbf{Frog} & \textbf{Horse} & \textbf{Ship} & \textbf{Truck} & \textbf{Avg}  \\ \hline
		VAE \cite{Kingma2014}      & 70.0     & 38.6       & 67.9 & 53.5 & 74.8 & 52.3 & 68.7 & 49.3  & 69.6 & 38.6  & 58.3 \\
		OCSVM \cite{Schlkopf1999SupportVM}    & 63.0     & 44.0       & 64.9 & 48.7 & 73.5 & 50.0 & 72.5 & 53.3  & 64.9 & 50.8  & 58.5 \\
		AnoGAN \cite{schlegl2017anogan}  & 67.1     & 54.7       & 52.9 & 54.5 & 65.1 & 60.3 & 58.5 & 62.5  & 75.8 & 66.5  & 61.8 \\
		PixelCNN \cite{van2016pixcnn} & 53.1     & 99.5       & 47.6 & 51.7 & 73.9 & 54.2 & 59.2 & 78.9  & 34.0 & 66.2  & 61.8 \\
		Deep-SVDD \cite{ruff18deepSvdd}   & 61.7     & 65.9       & 50.8 & 59.1 & 60.9 & 65.7 & 67.7 & 67.3  & 75.9 & 73.1  & 64.8 \\
		OCGAN \cite{perera2019ocgan}   & 75.7     & 53.1       & 64.0 & 62.0 & 72.3 & 62.0 & 72.3 & 57.5  & 82.0 & 55.4  & 65.6 \\
		Puzzle-AE \cite{DBLP:journals/corr/abs-2008-12959} & 78.9 & 78.0 & 69.9 & 54.8 & 75.4 & 66.0 & 74.7 & 73.3 & 83.3 & 69.9 & 72.4 \\
		DROCC \cite{goyal2020DROCC}   & 81.7     & 76.7       & 66.7 & 67.1 & 73.6 & 74.4 & 74.4 & 71.4  & 80.0 & 76.2  & 74.2 \\
		AnoNAGN \cite{DBLP:journals/corr/abs-2002-00522} & 96.2 & 63.8 & 72.5 & 64.3 & 87.3 & 63.8 & 88.3 & 58.4 & 93.5 & 64.5 & 75.01 \\
		GeoTrans \cite{Golan2018DeepAD}  & 74.7     & 95.7       & 78.1 & 72.4 & 87.8 & 87.8 & 83.4 & 95.5  & 93.3 & 91.3  & 86.0 \\
		PuzzleGeom \cite{Jezequel2021} & 75.1         & 96.3           & 84.8     & 74.2     & 91.1     & 89.9     & 88.7     &  95.5     &  94.7    &  91.9     & 88.2 \\
		SSD \cite{sehwag2021ssd} & 82.7 & 98.5 & 84.2 & 84.5 & 84.8 & 90.9 & 91.7 & 95.2 & 92.9 & 94.4 & 90.0 \\
		\hdashline
		Ours & 85.9 & 97.9 & 88.7 & 81.2 & 95.4 & 94.2 & 92.1 & 96.9 & 96.5 & 95.4 & \textbf{92.5}\\
		\tablebottom
	\end{tabular}
\end{table*}

%% file: figures/puzzle_n_piece_comp.pgf
\begingroup%
\makeatletter%
\begin{pgfpicture}%
\pgfpathrectangle{\pgfpointorigin}{\pgfqpoint{6.000000in}{2.000000in}}%
\pgfusepath{use as bounding box, clip}%
\begin{pgfscope}%
\pgfsetbuttcap%
\pgfsetmiterjoin%
\pgfsetlinewidth{0.000000pt}%
\definecolor{currentstroke}{rgb}{1.000000,1.000000,1.000000}%
\pgfsetstrokecolor{currentstroke}%
\pgfsetstrokeopacity{0.000000}%
\pgfsetdash{}{0pt}%
\pgfpathmoveto{\pgfqpoint{0.000000in}{0.000000in}}%
\pgfpathlineto{\pgfqpoint{6.000000in}{0.000000in}}%
\pgfpathlineto{\pgfqpoint{6.000000in}{2.000000in}}%
\pgfpathlineto{\pgfqpoint{0.000000in}{2.000000in}}%
\pgfpathclose%
\pgfusepath{}%
\end{pgfscope}%
\begin{pgfscope}%
\pgfsetbuttcap%
\pgfsetmiterjoin%
\definecolor{currentfill}{rgb}{1.000000,1.000000,1.000000}%
\pgfsetfillcolor{currentfill}%
\pgfsetlinewidth{0.000000pt}%
\definecolor{currentstroke}{rgb}{0.000000,0.000000,0.000000}%
\pgfsetstrokecolor{currentstroke}%
\pgfsetstrokeopacity{0.000000}%
\pgfsetdash{}{0pt}%
\pgfpathmoveto{\pgfqpoint{0.750000in}{0.250000in}}%
\pgfpathlineto{\pgfqpoint{5.400000in}{0.250000in}}%
\pgfpathlineto{\pgfqpoint{5.400000in}{1.760000in}}%
\pgfpathlineto{\pgfqpoint{0.750000in}{1.760000in}}%
\pgfpathclose%
\pgfusepath{fill}%
\end{pgfscope}%
\begin{pgfscope}%
\pgfpathrectangle{\pgfqpoint{0.750000in}{0.250000in}}{\pgfqpoint{4.650000in}{1.510000in}}%
\pgfusepath{clip}%
\pgfsetbuttcap%
\pgfsetroundjoin%
\definecolor{currentfill}{rgb}{0.121569,0.466667,0.705882}%
\pgfsetfillcolor{currentfill}%
\pgfsetlinewidth{1.016297pt}%
\definecolor{currentstroke}{rgb}{0.121569,0.466667,0.705882}%
\pgfsetstrokecolor{currentstroke}%
\pgfsetdash{}{0pt}%
\pgfsys@defobject{currentmarker}{\pgfqpoint{-0.023500in}{-0.023500in}}{\pgfqpoint{0.023500in}{0.023500in}}{%
\pgfpathmoveto{\pgfqpoint{-0.023500in}{0.000000in}}%
\pgfpathlineto{\pgfqpoint{0.023500in}{0.000000in}}%
\pgfpathmoveto{\pgfqpoint{0.000000in}{-0.023500in}}%
\pgfpathlineto{\pgfqpoint{0.000000in}{0.023500in}}%
\pgfusepath{stroke,fill}%
}%
\begin{pgfscope}%
\pgfsys@transformshift{1.191750in}{1.139463in}%
\pgfsys@useobject{currentmarker}{}%
\end{pgfscope}%
\begin{pgfscope}%
\pgfsys@transformshift{2.121750in}{1.060073in}%
\pgfsys@useobject{currentmarker}{}%
\end{pgfscope}%
\begin{pgfscope}%
\pgfsys@transformshift{3.051750in}{1.205728in}%
\pgfsys@useobject{currentmarker}{}%
\end{pgfscope}%
\begin{pgfscope}%
\pgfsys@transformshift{3.981750in}{1.040311in}%
\pgfsys@useobject{currentmarker}{}%
\end{pgfscope}%
\begin{pgfscope}%
\pgfsys@transformshift{4.911750in}{0.902635in}%
\pgfsys@useobject{currentmarker}{}%
\end{pgfscope}%
\end{pgfscope}%
\begin{pgfscope}%
\pgfpathrectangle{\pgfqpoint{0.750000in}{0.250000in}}{\pgfqpoint{4.650000in}{1.510000in}}%
\pgfusepath{clip}%
\pgfsetrectcap%
\pgfsetroundjoin%
\pgfsetlinewidth{1.355062pt}%
\definecolor{currentstroke}{rgb}{0.121569,0.466667,0.705882}%
\pgfsetstrokecolor{currentstroke}%
\pgfsetdash{}{0pt}%
\pgfpathmoveto{\pgfqpoint{1.191750in}{1.139463in}}%
\pgfpathlineto{\pgfqpoint{2.121750in}{1.060073in}}%
\pgfpathlineto{\pgfqpoint{3.051750in}{1.205728in}}%
\pgfpathlineto{\pgfqpoint{3.981750in}{1.040311in}}%
\pgfpathlineto{\pgfqpoint{4.911750in}{0.902635in}}%
\pgfusepath{stroke}%
\end{pgfscope}%
\begin{pgfscope}%
\pgfpathrectangle{\pgfqpoint{0.750000in}{0.250000in}}{\pgfqpoint{4.650000in}{1.510000in}}%
\pgfusepath{clip}%
\pgfsetrectcap%
\pgfsetroundjoin%
\pgfsetlinewidth{1.003750pt}%
\definecolor{currentstroke}{rgb}{0.121569,0.466667,0.705882}%
\pgfsetstrokecolor{currentstroke}%
\pgfsetdash{}{0pt}%
\pgfpathmoveto{\pgfqpoint{1.191750in}{0.743588in}}%
\pgfpathlineto{\pgfqpoint{1.191750in}{1.535339in}}%
\pgfusepath{stroke}%
\end{pgfscope}%
\begin{pgfscope}%
\pgfpathrectangle{\pgfqpoint{0.750000in}{0.250000in}}{\pgfqpoint{4.650000in}{1.510000in}}%
\pgfusepath{clip}%
\pgfsetrectcap%
\pgfsetroundjoin%
\pgfsetlinewidth{1.003750pt}%
\definecolor{currentstroke}{rgb}{0.121569,0.466667,0.705882}%
\pgfsetstrokecolor{currentstroke}%
\pgfsetdash{}{0pt}%
\pgfpathmoveto{\pgfqpoint{1.098750in}{0.743588in}}%
\pgfpathlineto{\pgfqpoint{1.284750in}{0.743588in}}%
\pgfusepath{stroke}%
\end{pgfscope}%
\begin{pgfscope}%
\pgfpathrectangle{\pgfqpoint{0.750000in}{0.250000in}}{\pgfqpoint{4.650000in}{1.510000in}}%
\pgfusepath{clip}%
\pgfsetrectcap%
\pgfsetroundjoin%
\pgfsetlinewidth{1.003750pt}%
\definecolor{currentstroke}{rgb}{0.121569,0.466667,0.705882}%
\pgfsetstrokecolor{currentstroke}%
\pgfsetdash{}{0pt}%
\pgfpathmoveto{\pgfqpoint{1.098750in}{1.535339in}}%
\pgfpathlineto{\pgfqpoint{1.284750in}{1.535339in}}%
\pgfusepath{stroke}%
\end{pgfscope}%
\begin{pgfscope}%
\pgfpathrectangle{\pgfqpoint{0.750000in}{0.250000in}}{\pgfqpoint{4.650000in}{1.510000in}}%
\pgfusepath{clip}%
\pgfsetrectcap%
\pgfsetroundjoin%
\pgfsetlinewidth{1.003750pt}%
\definecolor{currentstroke}{rgb}{0.121569,0.466667,0.705882}%
\pgfsetstrokecolor{currentstroke}%
\pgfsetdash{}{0pt}%
\pgfpathmoveto{\pgfqpoint{2.121750in}{0.594296in}}%
\pgfpathlineto{\pgfqpoint{2.121750in}{1.525851in}}%
\pgfusepath{stroke}%
\end{pgfscope}%
\begin{pgfscope}%
\pgfpathrectangle{\pgfqpoint{0.750000in}{0.250000in}}{\pgfqpoint{4.650000in}{1.510000in}}%
\pgfusepath{clip}%
\pgfsetrectcap%
\pgfsetroundjoin%
\pgfsetlinewidth{1.003750pt}%
\definecolor{currentstroke}{rgb}{0.121569,0.466667,0.705882}%
\pgfsetstrokecolor{currentstroke}%
\pgfsetdash{}{0pt}%
\pgfpathmoveto{\pgfqpoint{2.028750in}{0.594296in}}%
\pgfpathlineto{\pgfqpoint{2.214750in}{0.594296in}}%
\pgfusepath{stroke}%
\end{pgfscope}%
\begin{pgfscope}%
\pgfpathrectangle{\pgfqpoint{0.750000in}{0.250000in}}{\pgfqpoint{4.650000in}{1.510000in}}%
\pgfusepath{clip}%
\pgfsetrectcap%
\pgfsetroundjoin%
\pgfsetlinewidth{1.003750pt}%
\definecolor{currentstroke}{rgb}{0.121569,0.466667,0.705882}%
\pgfsetstrokecolor{currentstroke}%
\pgfsetdash{}{0pt}%
\pgfpathmoveto{\pgfqpoint{2.028750in}{1.525851in}}%
\pgfpathlineto{\pgfqpoint{2.214750in}{1.525851in}}%
\pgfusepath{stroke}%
\end{pgfscope}%
\begin{pgfscope}%
\pgfpathrectangle{\pgfqpoint{0.750000in}{0.250000in}}{\pgfqpoint{4.650000in}{1.510000in}}%
\pgfusepath{clip}%
\pgfsetrectcap%
\pgfsetroundjoin%
\pgfsetlinewidth{1.003750pt}%
\definecolor{currentstroke}{rgb}{0.121569,0.466667,0.705882}%
\pgfsetstrokecolor{currentstroke}%
\pgfsetdash{}{0pt}%
\pgfpathmoveto{\pgfqpoint{3.051750in}{0.720093in}}%
\pgfpathlineto{\pgfqpoint{3.051750in}{1.691364in}}%
\pgfusepath{stroke}%
\end{pgfscope}%
\begin{pgfscope}%
\pgfpathrectangle{\pgfqpoint{0.750000in}{0.250000in}}{\pgfqpoint{4.650000in}{1.510000in}}%
\pgfusepath{clip}%
\pgfsetrectcap%
\pgfsetroundjoin%
\pgfsetlinewidth{1.003750pt}%
\definecolor{currentstroke}{rgb}{0.121569,0.466667,0.705882}%
\pgfsetstrokecolor{currentstroke}%
\pgfsetdash{}{0pt}%
\pgfpathmoveto{\pgfqpoint{2.958750in}{0.720093in}}%
\pgfpathlineto{\pgfqpoint{3.144750in}{0.720093in}}%
\pgfusepath{stroke}%
\end{pgfscope}%
\begin{pgfscope}%
\pgfpathrectangle{\pgfqpoint{0.750000in}{0.250000in}}{\pgfqpoint{4.650000in}{1.510000in}}%
\pgfusepath{clip}%
\pgfsetrectcap%
\pgfsetroundjoin%
\pgfsetlinewidth{1.003750pt}%
\definecolor{currentstroke}{rgb}{0.121569,0.466667,0.705882}%
\pgfsetstrokecolor{currentstroke}%
\pgfsetdash{}{0pt}%
\pgfpathmoveto{\pgfqpoint{2.958750in}{1.691364in}}%
\pgfpathlineto{\pgfqpoint{3.144750in}{1.691364in}}%
\pgfusepath{stroke}%
\end{pgfscope}%
\begin{pgfscope}%
\pgfpathrectangle{\pgfqpoint{0.750000in}{0.250000in}}{\pgfqpoint{4.650000in}{1.510000in}}%
\pgfusepath{clip}%
\pgfsetrectcap%
\pgfsetroundjoin%
\pgfsetlinewidth{1.003750pt}%
\definecolor{currentstroke}{rgb}{0.121569,0.466667,0.705882}%
\pgfsetstrokecolor{currentstroke}%
\pgfsetdash{}{0pt}%
\pgfpathmoveto{\pgfqpoint{3.981750in}{0.494208in}}%
\pgfpathlineto{\pgfqpoint{3.981750in}{1.586414in}}%
\pgfusepath{stroke}%
\end{pgfscope}%
\begin{pgfscope}%
\pgfpathrectangle{\pgfqpoint{0.750000in}{0.250000in}}{\pgfqpoint{4.650000in}{1.510000in}}%
\pgfusepath{clip}%
\pgfsetrectcap%
\pgfsetroundjoin%
\pgfsetlinewidth{1.003750pt}%
\definecolor{currentstroke}{rgb}{0.121569,0.466667,0.705882}%
\pgfsetstrokecolor{currentstroke}%
\pgfsetdash{}{0pt}%
\pgfpathmoveto{\pgfqpoint{3.888750in}{0.494208in}}%
\pgfpathlineto{\pgfqpoint{4.074750in}{0.494208in}}%
\pgfusepath{stroke}%
\end{pgfscope}%
\begin{pgfscope}%
\pgfpathrectangle{\pgfqpoint{0.750000in}{0.250000in}}{\pgfqpoint{4.650000in}{1.510000in}}%
\pgfusepath{clip}%
\pgfsetrectcap%
\pgfsetroundjoin%
\pgfsetlinewidth{1.003750pt}%
\definecolor{currentstroke}{rgb}{0.121569,0.466667,0.705882}%
\pgfsetstrokecolor{currentstroke}%
\pgfsetdash{}{0pt}%
\pgfpathmoveto{\pgfqpoint{3.888750in}{1.586414in}}%
\pgfpathlineto{\pgfqpoint{4.074750in}{1.586414in}}%
\pgfusepath{stroke}%
\end{pgfscope}%
\begin{pgfscope}%
\pgfpathrectangle{\pgfqpoint{0.750000in}{0.250000in}}{\pgfqpoint{4.650000in}{1.510000in}}%
\pgfusepath{clip}%
\pgfsetrectcap%
\pgfsetroundjoin%
\pgfsetlinewidth{1.003750pt}%
\definecolor{currentstroke}{rgb}{0.121569,0.466667,0.705882}%
\pgfsetstrokecolor{currentstroke}%
\pgfsetdash{}{0pt}%
\pgfpathmoveto{\pgfqpoint{4.911750in}{0.318636in}}%
\pgfpathlineto{\pgfqpoint{4.911750in}{1.486634in}}%
\pgfusepath{stroke}%
\end{pgfscope}%
\begin{pgfscope}%
\pgfpathrectangle{\pgfqpoint{0.750000in}{0.250000in}}{\pgfqpoint{4.650000in}{1.510000in}}%
\pgfusepath{clip}%
\pgfsetrectcap%
\pgfsetroundjoin%
\pgfsetlinewidth{1.003750pt}%
\definecolor{currentstroke}{rgb}{0.121569,0.466667,0.705882}%
\pgfsetstrokecolor{currentstroke}%
\pgfsetdash{}{0pt}%
\pgfpathmoveto{\pgfqpoint{4.818750in}{0.318636in}}%
\pgfpathlineto{\pgfqpoint{5.004750in}{0.318636in}}%
\pgfusepath{stroke}%
\end{pgfscope}%
\begin{pgfscope}%
\pgfpathrectangle{\pgfqpoint{0.750000in}{0.250000in}}{\pgfqpoint{4.650000in}{1.510000in}}%
\pgfusepath{clip}%
\pgfsetrectcap%
\pgfsetroundjoin%
\pgfsetlinewidth{1.003750pt}%
\definecolor{currentstroke}{rgb}{0.121569,0.466667,0.705882}%
\pgfsetstrokecolor{currentstroke}%
\pgfsetdash{}{0pt}%
\pgfpathmoveto{\pgfqpoint{4.818750in}{1.486634in}}%
\pgfpathlineto{\pgfqpoint{5.004750in}{1.486634in}}%
\pgfusepath{stroke}%
\end{pgfscope}%
\begin{pgfscope}%
\pgfsetbuttcap%
\pgfsetroundjoin%
\definecolor{currentfill}{rgb}{0.000000,0.000000,0.000000}%
\pgfsetfillcolor{currentfill}%
\pgfsetlinewidth{0.803000pt}%
\definecolor{currentstroke}{rgb}{0.000000,0.000000,0.000000}%
\pgfsetstrokecolor{currentstroke}%
\pgfsetdash{}{0pt}%
\pgfsys@defobject{currentmarker}{\pgfqpoint{0.000000in}{-0.048611in}}{\pgfqpoint{0.000000in}{0.000000in}}{%
\pgfpathmoveto{\pgfqpoint{0.000000in}{0.000000in}}%
\pgfpathlineto{\pgfqpoint{0.000000in}{-0.048611in}}%
\pgfusepath{stroke,fill}%
}%
\begin{pgfscope}%
\pgfsys@transformshift{1.215000in}{0.250000in}%
\pgfsys@useobject{currentmarker}{}%
\end{pgfscope}%
\end{pgfscope}%
\begin{pgfscope}%
\definecolor{textcolor}{rgb}{0.000000,0.000000,0.000000}%
\pgfsetstrokecolor{textcolor}%
\pgfsetfillcolor{textcolor}%
\pgftext[x=1.215000in,y=0.152778in,,top]{\color{textcolor}\rmfamily\fontsize{10.000000}{12.000000}\selectfont 3x2}%
\end{pgfscope}%
\begin{pgfscope}%
\pgfsetbuttcap%
\pgfsetroundjoin%
\definecolor{currentfill}{rgb}{0.000000,0.000000,0.000000}%
\pgfsetfillcolor{currentfill}%
\pgfsetlinewidth{0.803000pt}%
\definecolor{currentstroke}{rgb}{0.000000,0.000000,0.000000}%
\pgfsetstrokecolor{currentstroke}%
\pgfsetdash{}{0pt}%
\pgfsys@defobject{currentmarker}{\pgfqpoint{0.000000in}{-0.048611in}}{\pgfqpoint{0.000000in}{0.000000in}}{%
\pgfpathmoveto{\pgfqpoint{0.000000in}{0.000000in}}%
\pgfpathlineto{\pgfqpoint{0.000000in}{-0.048611in}}%
\pgfusepath{stroke,fill}%
}%
\begin{pgfscope}%
\pgfsys@transformshift{2.145000in}{0.250000in}%
\pgfsys@useobject{currentmarker}{}%
\end{pgfscope}%
\end{pgfscope}%
\begin{pgfscope}%
\definecolor{textcolor}{rgb}{0.000000,0.000000,0.000000}%
\pgfsetstrokecolor{textcolor}%
\pgfsetfillcolor{textcolor}%
\pgftext[x=2.145000in,y=0.152778in,,top]{\color{textcolor}\rmfamily\fontsize{10.000000}{12.000000}\selectfont 4x2}%
\end{pgfscope}%
\begin{pgfscope}%
\pgfsetbuttcap%
\pgfsetroundjoin%
\definecolor{currentfill}{rgb}{0.000000,0.000000,0.000000}%
\pgfsetfillcolor{currentfill}%
\pgfsetlinewidth{0.803000pt}%
\definecolor{currentstroke}{rgb}{0.000000,0.000000,0.000000}%
\pgfsetstrokecolor{currentstroke}%
\pgfsetdash{}{0pt}%
\pgfsys@defobject{currentmarker}{\pgfqpoint{0.000000in}{-0.048611in}}{\pgfqpoint{0.000000in}{0.000000in}}{%
\pgfpathmoveto{\pgfqpoint{0.000000in}{0.000000in}}%
\pgfpathlineto{\pgfqpoint{0.000000in}{-0.048611in}}%
\pgfusepath{stroke,fill}%
}%
\begin{pgfscope}%
\pgfsys@transformshift{3.075000in}{0.250000in}%
\pgfsys@useobject{currentmarker}{}%
\end{pgfscope}%
\end{pgfscope}%
\begin{pgfscope}%
\definecolor{textcolor}{rgb}{0.000000,0.000000,0.000000}%
\pgfsetstrokecolor{textcolor}%
\pgfsetfillcolor{textcolor}%
\pgftext[x=3.075000in,y=0.152778in,,top]{\color{textcolor}\rmfamily\fontsize{10.000000}{12.000000}\selectfont 3x3}%
\end{pgfscope}%
\begin{pgfscope}%
\pgfsetbuttcap%
\pgfsetroundjoin%
\definecolor{currentfill}{rgb}{0.000000,0.000000,0.000000}%
\pgfsetfillcolor{currentfill}%
\pgfsetlinewidth{0.803000pt}%
\definecolor{currentstroke}{rgb}{0.000000,0.000000,0.000000}%
\pgfsetstrokecolor{currentstroke}%
\pgfsetdash{}{0pt}%
\pgfsys@defobject{currentmarker}{\pgfqpoint{0.000000in}{-0.048611in}}{\pgfqpoint{0.000000in}{0.000000in}}{%
\pgfpathmoveto{\pgfqpoint{0.000000in}{0.000000in}}%
\pgfpathlineto{\pgfqpoint{0.000000in}{-0.048611in}}%
\pgfusepath{stroke,fill}%
}%
\begin{pgfscope}%
\pgfsys@transformshift{4.005000in}{0.250000in}%
\pgfsys@useobject{currentmarker}{}%
\end{pgfscope}%
\end{pgfscope}%
\begin{pgfscope}%
\definecolor{textcolor}{rgb}{0.000000,0.000000,0.000000}%
\pgfsetstrokecolor{textcolor}%
\pgfsetfillcolor{textcolor}%
\pgftext[x=4.005000in,y=0.152778in,,top]{\color{textcolor}\rmfamily\fontsize{10.000000}{12.000000}\selectfont 4x3}%
\end{pgfscope}%
\begin{pgfscope}%
\pgfsetbuttcap%
\pgfsetroundjoin%
\definecolor{currentfill}{rgb}{0.000000,0.000000,0.000000}%
\pgfsetfillcolor{currentfill}%
\pgfsetlinewidth{0.803000pt}%
\definecolor{currentstroke}{rgb}{0.000000,0.000000,0.000000}%
\pgfsetstrokecolor{currentstroke}%
\pgfsetdash{}{0pt}%
\pgfsys@defobject{currentmarker}{\pgfqpoint{0.000000in}{-0.048611in}}{\pgfqpoint{0.000000in}{0.000000in}}{%
\pgfpathmoveto{\pgfqpoint{0.000000in}{0.000000in}}%
\pgfpathlineto{\pgfqpoint{0.000000in}{-0.048611in}}%
\pgfusepath{stroke,fill}%
}%
\begin{pgfscope}%
\pgfsys@transformshift{4.935000in}{0.250000in}%
\pgfsys@useobject{currentmarker}{}%
\end{pgfscope}%
\end{pgfscope}%
\begin{pgfscope}%
\definecolor{textcolor}{rgb}{0.000000,0.000000,0.000000}%
\pgfsetstrokecolor{textcolor}%
\pgfsetfillcolor{textcolor}%
\pgftext[x=4.935000in,y=0.152778in,,top]{\color{textcolor}\rmfamily\fontsize{10.000000}{12.000000}\selectfont 4x4}%
\end{pgfscope}%
\begin{pgfscope}%
\definecolor{textcolor}{rgb}{0.000000,0.000000,0.000000}%
\pgfsetstrokecolor{textcolor}%
\pgfsetfillcolor{textcolor}%
\pgftext[x=3.075000in,y=-0.026234in,,top]{\color{textcolor}\rmfamily\fontsize{10.000000}{12.000000}\selectfont n}%
\end{pgfscope}%
\begin{pgfscope}%
\pgfpathrectangle{\pgfqpoint{0.750000in}{0.250000in}}{\pgfqpoint{4.650000in}{1.510000in}}%
\pgfusepath{clip}%
\pgfsetbuttcap%
\pgfsetroundjoin%
\pgfsetlinewidth{0.803000pt}%
\definecolor{currentstroke}{rgb}{0.690196,0.690196,0.690196}%
\pgfsetstrokecolor{currentstroke}%
\pgfsetdash{{2.960000pt}{1.280000pt}}{0.000000pt}%
\pgfpathmoveto{\pgfqpoint{0.750000in}{0.407650in}}%
\pgfpathlineto{\pgfqpoint{5.400000in}{0.407650in}}%
\pgfusepath{stroke}%
\end{pgfscope}%
\begin{pgfscope}%
\pgfsetbuttcap%
\pgfsetroundjoin%
\definecolor{currentfill}{rgb}{0.000000,0.000000,0.000000}%
\pgfsetfillcolor{currentfill}%
\pgfsetlinewidth{0.803000pt}%
\definecolor{currentstroke}{rgb}{0.000000,0.000000,0.000000}%
\pgfsetstrokecolor{currentstroke}%
\pgfsetdash{}{0pt}%
\pgfsys@defobject{currentmarker}{\pgfqpoint{-0.048611in}{0.000000in}}{\pgfqpoint{-0.000000in}{0.000000in}}{%
\pgfpathmoveto{\pgfqpoint{-0.000000in}{0.000000in}}%
\pgfpathlineto{\pgfqpoint{-0.048611in}{0.000000in}}%
\pgfusepath{stroke,fill}%
}%
\begin{pgfscope}%
\pgfsys@transformshift{0.750000in}{0.407650in}%
\pgfsys@useobject{currentmarker}{}%
\end{pgfscope}%
\end{pgfscope}%
\begin{pgfscope}%
\definecolor{textcolor}{rgb}{0.000000,0.000000,0.000000}%
\pgfsetstrokecolor{textcolor}%
\pgfsetfillcolor{textcolor}%
\pgftext[x=0.513888in, y=0.359425in, left, base]{\color{textcolor}\rmfamily\fontsize{10.000000}{12.000000}\selectfont \(\displaystyle {70}\)}%
\end{pgfscope}%
\begin{pgfscope}%
\pgfpathrectangle{\pgfqpoint{0.750000in}{0.250000in}}{\pgfqpoint{4.650000in}{1.510000in}}%
\pgfusepath{clip}%
\pgfsetbuttcap%
\pgfsetroundjoin%
\pgfsetlinewidth{0.803000pt}%
\definecolor{currentstroke}{rgb}{0.690196,0.690196,0.690196}%
\pgfsetstrokecolor{currentstroke}%
\pgfsetdash{{2.960000pt}{1.280000pt}}{0.000000pt}%
\pgfpathmoveto{\pgfqpoint{0.750000in}{0.916561in}}%
\pgfpathlineto{\pgfqpoint{5.400000in}{0.916561in}}%
\pgfusepath{stroke}%
\end{pgfscope}%
\begin{pgfscope}%
\pgfsetbuttcap%
\pgfsetroundjoin%
\definecolor{currentfill}{rgb}{0.000000,0.000000,0.000000}%
\pgfsetfillcolor{currentfill}%
\pgfsetlinewidth{0.803000pt}%
\definecolor{currentstroke}{rgb}{0.000000,0.000000,0.000000}%
\pgfsetstrokecolor{currentstroke}%
\pgfsetdash{}{0pt}%
\pgfsys@defobject{currentmarker}{\pgfqpoint{-0.048611in}{0.000000in}}{\pgfqpoint{-0.000000in}{0.000000in}}{%
\pgfpathmoveto{\pgfqpoint{-0.000000in}{0.000000in}}%
\pgfpathlineto{\pgfqpoint{-0.048611in}{0.000000in}}%
\pgfusepath{stroke,fill}%
}%
\begin{pgfscope}%
\pgfsys@transformshift{0.750000in}{0.916561in}%
\pgfsys@useobject{currentmarker}{}%
\end{pgfscope}%
\end{pgfscope}%
\begin{pgfscope}%
\definecolor{textcolor}{rgb}{0.000000,0.000000,0.000000}%
\pgfsetstrokecolor{textcolor}%
\pgfsetfillcolor{textcolor}%
\pgftext[x=0.513888in, y=0.868335in, left, base]{\color{textcolor}\rmfamily\fontsize{10.000000}{12.000000}\selectfont \(\displaystyle {80}\)}%
\end{pgfscope}%
\begin{pgfscope}%
\pgfpathrectangle{\pgfqpoint{0.750000in}{0.250000in}}{\pgfqpoint{4.650000in}{1.510000in}}%
\pgfusepath{clip}%
\pgfsetbuttcap%
\pgfsetroundjoin%
\pgfsetlinewidth{0.803000pt}%
\definecolor{currentstroke}{rgb}{0.690196,0.690196,0.690196}%
\pgfsetstrokecolor{currentstroke}%
\pgfsetdash{{2.960000pt}{1.280000pt}}{0.000000pt}%
\pgfpathmoveto{\pgfqpoint{0.750000in}{1.425471in}}%
\pgfpathlineto{\pgfqpoint{5.400000in}{1.425471in}}%
\pgfusepath{stroke}%
\end{pgfscope}%
\begin{pgfscope}%
\pgfsetbuttcap%
\pgfsetroundjoin%
\definecolor{currentfill}{rgb}{0.000000,0.000000,0.000000}%
\pgfsetfillcolor{currentfill}%
\pgfsetlinewidth{0.803000pt}%
\definecolor{currentstroke}{rgb}{0.000000,0.000000,0.000000}%
\pgfsetstrokecolor{currentstroke}%
\pgfsetdash{}{0pt}%
\pgfsys@defobject{currentmarker}{\pgfqpoint{-0.048611in}{0.000000in}}{\pgfqpoint{-0.000000in}{0.000000in}}{%
\pgfpathmoveto{\pgfqpoint{-0.000000in}{0.000000in}}%
\pgfpathlineto{\pgfqpoint{-0.048611in}{0.000000in}}%
\pgfusepath{stroke,fill}%
}%
\begin{pgfscope}%
\pgfsys@transformshift{0.750000in}{1.425471in}%
\pgfsys@useobject{currentmarker}{}%
\end{pgfscope}%
\end{pgfscope}%
\begin{pgfscope}%
\definecolor{textcolor}{rgb}{0.000000,0.000000,0.000000}%
\pgfsetstrokecolor{textcolor}%
\pgfsetfillcolor{textcolor}%
\pgftext[x=0.513888in, y=1.377246in, left, base]{\color{textcolor}\rmfamily\fontsize{10.000000}{12.000000}\selectfont \(\displaystyle {90}\)}%
\end{pgfscope}%
\begin{pgfscope}%
\definecolor{textcolor}{rgb}{0.000000,0.000000,0.000000}%
\pgfsetstrokecolor{textcolor}%
\pgfsetfillcolor{textcolor}%
\pgftext[x=0.458333in,y=1.005000in,,bottom,rotate=90.000000]{\color{textcolor}\rmfamily\fontsize{10.000000}{12.000000}\selectfont AUC}%
\end{pgfscope}%
\begin{pgfscope}%
\pgfpathrectangle{\pgfqpoint{0.750000in}{0.250000in}}{\pgfqpoint{4.650000in}{1.510000in}}%
\pgfusepath{clip}%
\pgfsetbuttcap%
\pgfsetroundjoin%
\definecolor{currentfill}{rgb}{1.000000,0.498039,0.054902}%
\pgfsetfillcolor{currentfill}%
\pgfsetlinewidth{1.016297pt}%
\definecolor{currentstroke}{rgb}{1.000000,0.498039,0.054902}%
\pgfsetstrokecolor{currentstroke}%
\pgfsetdash{}{0pt}%
\pgfsys@defobject{currentmarker}{\pgfqpoint{-0.023500in}{-0.023500in}}{\pgfqpoint{0.023500in}{0.023500in}}{%
\pgfpathmoveto{\pgfqpoint{-0.023500in}{0.000000in}}%
\pgfpathlineto{\pgfqpoint{0.023500in}{0.000000in}}%
\pgfpathmoveto{\pgfqpoint{0.000000in}{-0.023500in}}%
\pgfpathlineto{\pgfqpoint{0.000000in}{0.023500in}}%
\pgfusepath{stroke,fill}%
}%
\begin{pgfscope}%
\pgfsys@transformshift{1.238250in}{1.154061in}%
\pgfsys@useobject{currentmarker}{}%
\end{pgfscope}%
\begin{pgfscope}%
\pgfsys@transformshift{2.168250in}{1.069234in}%
\pgfsys@useobject{currentmarker}{}%
\end{pgfscope}%
\begin{pgfscope}%
\pgfsys@transformshift{3.098250in}{1.227337in}%
\pgfsys@useobject{currentmarker}{}%
\end{pgfscope}%
\begin{pgfscope}%
\pgfsys@transformshift{4.028250in}{1.184247in}%
\pgfsys@useobject{currentmarker}{}%
\end{pgfscope}%
\begin{pgfscope}%
\pgfsys@transformshift{4.958250in}{1.026994in}%
\pgfsys@useobject{currentmarker}{}%
\end{pgfscope}%
\end{pgfscope}%
\begin{pgfscope}%
\pgfpathrectangle{\pgfqpoint{0.750000in}{0.250000in}}{\pgfqpoint{4.650000in}{1.510000in}}%
\pgfusepath{clip}%
\pgfsetrectcap%
\pgfsetroundjoin%
\pgfsetlinewidth{1.355062pt}%
\definecolor{currentstroke}{rgb}{1.000000,0.498039,0.054902}%
\pgfsetstrokecolor{currentstroke}%
\pgfsetdash{}{0pt}%
\pgfpathmoveto{\pgfqpoint{1.238250in}{1.154061in}}%
\pgfpathlineto{\pgfqpoint{2.168250in}{1.069234in}}%
\pgfpathlineto{\pgfqpoint{3.098250in}{1.227337in}}%
\pgfpathlineto{\pgfqpoint{4.028250in}{1.184247in}}%
\pgfpathlineto{\pgfqpoint{4.958250in}{1.026994in}}%
\pgfusepath{stroke}%
\end{pgfscope}%
\begin{pgfscope}%
\pgfpathrectangle{\pgfqpoint{0.750000in}{0.250000in}}{\pgfqpoint{4.650000in}{1.510000in}}%
\pgfusepath{clip}%
\pgfsetrectcap%
\pgfsetroundjoin%
\pgfsetlinewidth{1.003750pt}%
\definecolor{currentstroke}{rgb}{1.000000,0.498039,0.054902}%
\pgfsetstrokecolor{currentstroke}%
\pgfsetdash{}{0pt}%
\pgfpathmoveto{\pgfqpoint{1.238250in}{0.751909in}}%
\pgfpathlineto{\pgfqpoint{1.238250in}{1.556214in}}%
\pgfusepath{stroke}%
\end{pgfscope}%
\begin{pgfscope}%
\pgfpathrectangle{\pgfqpoint{0.750000in}{0.250000in}}{\pgfqpoint{4.650000in}{1.510000in}}%
\pgfusepath{clip}%
\pgfsetrectcap%
\pgfsetroundjoin%
\pgfsetlinewidth{1.003750pt}%
\definecolor{currentstroke}{rgb}{1.000000,0.498039,0.054902}%
\pgfsetstrokecolor{currentstroke}%
\pgfsetdash{}{0pt}%
\pgfpathmoveto{\pgfqpoint{1.145250in}{0.751909in}}%
\pgfpathlineto{\pgfqpoint{1.331250in}{0.751909in}}%
\pgfusepath{stroke}%
\end{pgfscope}%
\begin{pgfscope}%
\pgfpathrectangle{\pgfqpoint{0.750000in}{0.250000in}}{\pgfqpoint{4.650000in}{1.510000in}}%
\pgfusepath{clip}%
\pgfsetrectcap%
\pgfsetroundjoin%
\pgfsetlinewidth{1.003750pt}%
\definecolor{currentstroke}{rgb}{1.000000,0.498039,0.054902}%
\pgfsetstrokecolor{currentstroke}%
\pgfsetdash{}{0pt}%
\pgfpathmoveto{\pgfqpoint{1.145250in}{1.556214in}}%
\pgfpathlineto{\pgfqpoint{1.331250in}{1.556214in}}%
\pgfusepath{stroke}%
\end{pgfscope}%
\begin{pgfscope}%
\pgfpathrectangle{\pgfqpoint{0.750000in}{0.250000in}}{\pgfqpoint{4.650000in}{1.510000in}}%
\pgfusepath{clip}%
\pgfsetrectcap%
\pgfsetroundjoin%
\pgfsetlinewidth{1.003750pt}%
\definecolor{currentstroke}{rgb}{1.000000,0.498039,0.054902}%
\pgfsetstrokecolor{currentstroke}%
\pgfsetdash{}{0pt}%
\pgfpathmoveto{\pgfqpoint{2.168250in}{0.645281in}}%
\pgfpathlineto{\pgfqpoint{2.168250in}{1.493187in}}%
\pgfusepath{stroke}%
\end{pgfscope}%
\begin{pgfscope}%
\pgfpathrectangle{\pgfqpoint{0.750000in}{0.250000in}}{\pgfqpoint{4.650000in}{1.510000in}}%
\pgfusepath{clip}%
\pgfsetrectcap%
\pgfsetroundjoin%
\pgfsetlinewidth{1.003750pt}%
\definecolor{currentstroke}{rgb}{1.000000,0.498039,0.054902}%
\pgfsetstrokecolor{currentstroke}%
\pgfsetdash{}{0pt}%
\pgfpathmoveto{\pgfqpoint{2.075250in}{0.645281in}}%
\pgfpathlineto{\pgfqpoint{2.261250in}{0.645281in}}%
\pgfusepath{stroke}%
\end{pgfscope}%
\begin{pgfscope}%
\pgfpathrectangle{\pgfqpoint{0.750000in}{0.250000in}}{\pgfqpoint{4.650000in}{1.510000in}}%
\pgfusepath{clip}%
\pgfsetrectcap%
\pgfsetroundjoin%
\pgfsetlinewidth{1.003750pt}%
\definecolor{currentstroke}{rgb}{1.000000,0.498039,0.054902}%
\pgfsetstrokecolor{currentstroke}%
\pgfsetdash{}{0pt}%
\pgfpathmoveto{\pgfqpoint{2.075250in}{1.493187in}}%
\pgfpathlineto{\pgfqpoint{2.261250in}{1.493187in}}%
\pgfusepath{stroke}%
\end{pgfscope}%
\begin{pgfscope}%
\pgfpathrectangle{\pgfqpoint{0.750000in}{0.250000in}}{\pgfqpoint{4.650000in}{1.510000in}}%
\pgfusepath{clip}%
\pgfsetrectcap%
\pgfsetroundjoin%
\pgfsetlinewidth{1.003750pt}%
\definecolor{currentstroke}{rgb}{1.000000,0.498039,0.054902}%
\pgfsetstrokecolor{currentstroke}%
\pgfsetdash{}{0pt}%
\pgfpathmoveto{\pgfqpoint{3.098250in}{0.805309in}}%
\pgfpathlineto{\pgfqpoint{3.098250in}{1.649366in}}%
\pgfusepath{stroke}%
\end{pgfscope}%
\begin{pgfscope}%
\pgfpathrectangle{\pgfqpoint{0.750000in}{0.250000in}}{\pgfqpoint{4.650000in}{1.510000in}}%
\pgfusepath{clip}%
\pgfsetrectcap%
\pgfsetroundjoin%
\pgfsetlinewidth{1.003750pt}%
\definecolor{currentstroke}{rgb}{1.000000,0.498039,0.054902}%
\pgfsetstrokecolor{currentstroke}%
\pgfsetdash{}{0pt}%
\pgfpathmoveto{\pgfqpoint{3.005250in}{0.805309in}}%
\pgfpathlineto{\pgfqpoint{3.191250in}{0.805309in}}%
\pgfusepath{stroke}%
\end{pgfscope}%
\begin{pgfscope}%
\pgfpathrectangle{\pgfqpoint{0.750000in}{0.250000in}}{\pgfqpoint{4.650000in}{1.510000in}}%
\pgfusepath{clip}%
\pgfsetrectcap%
\pgfsetroundjoin%
\pgfsetlinewidth{1.003750pt}%
\definecolor{currentstroke}{rgb}{1.000000,0.498039,0.054902}%
\pgfsetstrokecolor{currentstroke}%
\pgfsetdash{}{0pt}%
\pgfpathmoveto{\pgfqpoint{3.005250in}{1.649366in}}%
\pgfpathlineto{\pgfqpoint{3.191250in}{1.649366in}}%
\pgfusepath{stroke}%
\end{pgfscope}%
\begin{pgfscope}%
\pgfpathrectangle{\pgfqpoint{0.750000in}{0.250000in}}{\pgfqpoint{4.650000in}{1.510000in}}%
\pgfusepath{clip}%
\pgfsetrectcap%
\pgfsetroundjoin%
\pgfsetlinewidth{1.003750pt}%
\definecolor{currentstroke}{rgb}{1.000000,0.498039,0.054902}%
\pgfsetstrokecolor{currentstroke}%
\pgfsetdash{}{0pt}%
\pgfpathmoveto{\pgfqpoint{4.028250in}{0.705600in}}%
\pgfpathlineto{\pgfqpoint{4.028250in}{1.662895in}}%
\pgfusepath{stroke}%
\end{pgfscope}%
\begin{pgfscope}%
\pgfpathrectangle{\pgfqpoint{0.750000in}{0.250000in}}{\pgfqpoint{4.650000in}{1.510000in}}%
\pgfusepath{clip}%
\pgfsetrectcap%
\pgfsetroundjoin%
\pgfsetlinewidth{1.003750pt}%
\definecolor{currentstroke}{rgb}{1.000000,0.498039,0.054902}%
\pgfsetstrokecolor{currentstroke}%
\pgfsetdash{}{0pt}%
\pgfpathmoveto{\pgfqpoint{3.935250in}{0.705600in}}%
\pgfpathlineto{\pgfqpoint{4.121250in}{0.705600in}}%
\pgfusepath{stroke}%
\end{pgfscope}%
\begin{pgfscope}%
\pgfpathrectangle{\pgfqpoint{0.750000in}{0.250000in}}{\pgfqpoint{4.650000in}{1.510000in}}%
\pgfusepath{clip}%
\pgfsetrectcap%
\pgfsetroundjoin%
\pgfsetlinewidth{1.003750pt}%
\definecolor{currentstroke}{rgb}{1.000000,0.498039,0.054902}%
\pgfsetstrokecolor{currentstroke}%
\pgfsetdash{}{0pt}%
\pgfpathmoveto{\pgfqpoint{3.935250in}{1.662895in}}%
\pgfpathlineto{\pgfqpoint{4.121250in}{1.662895in}}%
\pgfusepath{stroke}%
\end{pgfscope}%
\begin{pgfscope}%
\pgfpathrectangle{\pgfqpoint{0.750000in}{0.250000in}}{\pgfqpoint{4.650000in}{1.510000in}}%
\pgfusepath{clip}%
\pgfsetrectcap%
\pgfsetroundjoin%
\pgfsetlinewidth{1.003750pt}%
\definecolor{currentstroke}{rgb}{1.000000,0.498039,0.054902}%
\pgfsetstrokecolor{currentstroke}%
\pgfsetdash{}{0pt}%
\pgfpathmoveto{\pgfqpoint{4.958250in}{0.473229in}}%
\pgfpathlineto{\pgfqpoint{4.958250in}{1.580759in}}%
\pgfusepath{stroke}%
\end{pgfscope}%
\begin{pgfscope}%
\pgfpathrectangle{\pgfqpoint{0.750000in}{0.250000in}}{\pgfqpoint{4.650000in}{1.510000in}}%
\pgfusepath{clip}%
\pgfsetrectcap%
\pgfsetroundjoin%
\pgfsetlinewidth{1.003750pt}%
\definecolor{currentstroke}{rgb}{1.000000,0.498039,0.054902}%
\pgfsetstrokecolor{currentstroke}%
\pgfsetdash{}{0pt}%
\pgfpathmoveto{\pgfqpoint{4.865250in}{0.473229in}}%
\pgfpathlineto{\pgfqpoint{5.051250in}{0.473229in}}%
\pgfusepath{stroke}%
\end{pgfscope}%
\begin{pgfscope}%
\pgfpathrectangle{\pgfqpoint{0.750000in}{0.250000in}}{\pgfqpoint{4.650000in}{1.510000in}}%
\pgfusepath{clip}%
\pgfsetrectcap%
\pgfsetroundjoin%
\pgfsetlinewidth{1.003750pt}%
\definecolor{currentstroke}{rgb}{1.000000,0.498039,0.054902}%
\pgfsetstrokecolor{currentstroke}%
\pgfsetdash{}{0pt}%
\pgfpathmoveto{\pgfqpoint{4.865250in}{1.580759in}}%
\pgfpathlineto{\pgfqpoint{5.051250in}{1.580759in}}%
\pgfusepath{stroke}%
\end{pgfscope}%
\begin{pgfscope}%
\pgfsetrectcap%
\pgfsetmiterjoin%
\pgfsetlinewidth{0.803000pt}%
\definecolor{currentstroke}{rgb}{0.000000,0.000000,0.000000}%
\pgfsetstrokecolor{currentstroke}%
\pgfsetdash{}{0pt}%
\pgfpathmoveto{\pgfqpoint{0.750000in}{0.250000in}}%
\pgfpathlineto{\pgfqpoint{0.750000in}{1.760000in}}%
\pgfusepath{stroke}%
\end{pgfscope}%
\begin{pgfscope}%
\pgfsetrectcap%
\pgfsetmiterjoin%
\pgfsetlinewidth{0.803000pt}%
\definecolor{currentstroke}{rgb}{0.000000,0.000000,0.000000}%
\pgfsetstrokecolor{currentstroke}%
\pgfsetdash{}{0pt}%
\pgfpathmoveto{\pgfqpoint{5.400000in}{0.250000in}}%
\pgfpathlineto{\pgfqpoint{5.400000in}{1.760000in}}%
\pgfusepath{stroke}%
\end{pgfscope}%
\begin{pgfscope}%
\pgfsetrectcap%
\pgfsetmiterjoin%
\pgfsetlinewidth{0.803000pt}%
\definecolor{currentstroke}{rgb}{0.000000,0.000000,0.000000}%
\pgfsetstrokecolor{currentstroke}%
\pgfsetdash{}{0pt}%
\pgfpathmoveto{\pgfqpoint{0.750000in}{0.250000in}}%
\pgfpathlineto{\pgfqpoint{5.400000in}{0.250000in}}%
\pgfusepath{stroke}%
\end{pgfscope}%
\begin{pgfscope}%
\pgfsetrectcap%
\pgfsetmiterjoin%
\pgfsetlinewidth{0.803000pt}%
\definecolor{currentstroke}{rgb}{0.000000,0.000000,0.000000}%
\pgfsetstrokecolor{currentstroke}%
\pgfsetdash{}{0pt}%
\pgfpathmoveto{\pgfqpoint{0.750000in}{1.760000in}}%
\pgfpathlineto{\pgfqpoint{5.400000in}{1.760000in}}%
\pgfusepath{stroke}%
\end{pgfscope}%
\begin{pgfscope}%
\pgfsetbuttcap%
\pgfsetmiterjoin%
\definecolor{currentfill}{rgb}{1.000000,1.000000,1.000000}%
\pgfsetfillcolor{currentfill}%
\pgfsetfillopacity{0.800000}%
\pgfsetlinewidth{1.003750pt}%
\definecolor{currentstroke}{rgb}{0.800000,0.800000,0.800000}%
\pgfsetstrokecolor{currentstroke}%
\pgfsetstrokeopacity{0.800000}%
\pgfsetdash{}{0pt}%
\pgfpathmoveto{\pgfqpoint{4.719444in}{0.694043in}}%
\pgfpathlineto{\pgfqpoint{5.302778in}{0.694043in}}%
\pgfpathquadraticcurveto{\pgfqpoint{5.330556in}{0.694043in}}{\pgfqpoint{5.330556in}{0.721821in}}%
\pgfpathlineto{\pgfqpoint{5.330556in}{1.288179in}}%
\pgfpathquadraticcurveto{\pgfqpoint{5.330556in}{1.315957in}}{\pgfqpoint{5.302778in}{1.315957in}}%
\pgfpathlineto{\pgfqpoint{4.719444in}{1.315957in}}%
\pgfpathquadraticcurveto{\pgfqpoint{4.691666in}{1.315957in}}{\pgfqpoint{4.691666in}{1.288179in}}%
\pgfpathlineto{\pgfqpoint{4.691666in}{0.721821in}}%
\pgfpathquadraticcurveto{\pgfqpoint{4.691666in}{0.694043in}}{\pgfqpoint{4.719444in}{0.694043in}}%
\pgfpathclose%
\pgfusepath{stroke,fill}%
\end{pgfscope}%
\begin{pgfscope}%
\definecolor{textcolor}{rgb}{0.000000,0.000000,0.000000}%
\pgfsetstrokecolor{textcolor}%
\pgfsetfillcolor{textcolor}%
\pgftext[x=4.974460in,y=1.163951in,left,base]{\color{textcolor}\rmfamily\fontsize{10.000000}{12.000000}\selectfont $\nrstest$}%
\end{pgfscope}%
\begin{pgfscope}%
\pgfsetbuttcap%
\pgfsetroundjoin%
\definecolor{currentfill}{rgb}{0.121569,0.466667,0.705882}%
\pgfsetfillcolor{currentfill}%
\pgfsetlinewidth{1.016297pt}%
\definecolor{currentstroke}{rgb}{0.121569,0.466667,0.705882}%
\pgfsetstrokecolor{currentstroke}%
\pgfsetdash{}{0pt}%
\pgfsys@defobject{currentmarker}{\pgfqpoint{-0.023500in}{-0.023500in}}{\pgfqpoint{0.023500in}{0.023500in}}{%
\pgfpathmoveto{\pgfqpoint{-0.023500in}{0.000000in}}%
\pgfpathlineto{\pgfqpoint{0.023500in}{0.000000in}}%
\pgfpathmoveto{\pgfqpoint{0.000000in}{-0.023500in}}%
\pgfpathlineto{\pgfqpoint{0.000000in}{0.023500in}}%
\pgfusepath{stroke,fill}%
}%
\begin{pgfscope}%
\pgfsys@transformshift{4.886111in}{1.006736in}%
\pgfsys@useobject{currentmarker}{}%
\end{pgfscope}%
\end{pgfscope}%
\begin{pgfscope}%
\definecolor{textcolor}{rgb}{0.000000,0.000000,0.000000}%
\pgfsetstrokecolor{textcolor}%
\pgfsetfillcolor{textcolor}%
\pgftext[x=5.136111in,y=0.970278in,left,base]{\color{textcolor}\rmfamily\fontsize{10.000000}{12.000000}\selectfont 9}%
\end{pgfscope}%
\begin{pgfscope}%
\pgfsetbuttcap%
\pgfsetroundjoin%
\definecolor{currentfill}{rgb}{1.000000,0.498039,0.054902}%
\pgfsetfillcolor{currentfill}%
\pgfsetlinewidth{1.016297pt}%
\definecolor{currentstroke}{rgb}{1.000000,0.498039,0.054902}%
\pgfsetstrokecolor{currentstroke}%
\pgfsetdash{}{0pt}%
\pgfsys@defobject{currentmarker}{\pgfqpoint{-0.023500in}{-0.023500in}}{\pgfqpoint{0.023500in}{0.023500in}}{%
\pgfpathmoveto{\pgfqpoint{-0.023500in}{0.000000in}}%
\pgfpathlineto{\pgfqpoint{0.023500in}{0.000000in}}%
\pgfpathmoveto{\pgfqpoint{0.000000in}{-0.023500in}}%
\pgfpathlineto{\pgfqpoint{0.000000in}{0.023500in}}%
\pgfusepath{stroke,fill}%
}%
\begin{pgfscope}%
\pgfsys@transformshift{4.886111in}{0.813063in}%
\pgfsys@useobject{currentmarker}{}%
\end{pgfscope}%
\end{pgfscope}%
\begin{pgfscope}%
\definecolor{textcolor}{rgb}{0.000000,0.000000,0.000000}%
\pgfsetstrokecolor{textcolor}%
\pgfsetfillcolor{textcolor}%
\pgftext[x=5.136111in,y=0.776605in,left,base]{\color{textcolor}\rmfamily\fontsize{10.000000}{12.000000}\selectfont 18}%
\end{pgfscope}%
\end{pgfpicture}%
\makeatother%
\endgroup%

%% file: figures/hue_rot_comp_auc_plot.pgf
\begingroup%
\makeatletter%
\begin{pgfpicture}%
\pgfpathrectangle{\pgfpointorigin}{\pgfqpoint{6.000000in}{2.000000in}}%
\pgfusepath{use as bounding box, clip}%
\begin{pgfscope}%
\pgfsetbuttcap%
\pgfsetmiterjoin%
\pgfsetlinewidth{0.000000pt}%
\definecolor{currentstroke}{rgb}{1.000000,1.000000,1.000000}%
\pgfsetstrokecolor{currentstroke}%
\pgfsetstrokeopacity{0.000000}%
\pgfsetdash{}{0pt}%
\pgfpathmoveto{\pgfqpoint{0.000000in}{0.000000in}}%
\pgfpathlineto{\pgfqpoint{6.000000in}{0.000000in}}%
\pgfpathlineto{\pgfqpoint{6.000000in}{2.000000in}}%
\pgfpathlineto{\pgfqpoint{0.000000in}{2.000000in}}%
\pgfpathclose%
\pgfusepath{}%
\end{pgfscope}%
\begin{pgfscope}%
\pgfsetbuttcap%
\pgfsetmiterjoin%
\definecolor{currentfill}{rgb}{1.000000,1.000000,1.000000}%
\pgfsetfillcolor{currentfill}%
\pgfsetlinewidth{0.000000pt}%
\definecolor{currentstroke}{rgb}{0.000000,0.000000,0.000000}%
\pgfsetstrokecolor{currentstroke}%
\pgfsetstrokeopacity{0.000000}%
\pgfsetdash{}{0pt}%
\pgfpathmoveto{\pgfqpoint{0.750000in}{0.250000in}}%
\pgfpathlineto{\pgfqpoint{5.400000in}{0.250000in}}%
\pgfpathlineto{\pgfqpoint{5.400000in}{1.760000in}}%
\pgfpathlineto{\pgfqpoint{0.750000in}{1.760000in}}%
\pgfpathclose%
\pgfusepath{fill}%
\end{pgfscope}%
\begin{pgfscope}%
\pgfpathrectangle{\pgfqpoint{0.750000in}{0.250000in}}{\pgfqpoint{4.650000in}{1.510000in}}%
\pgfusepath{clip}%
\pgfsetbuttcap%
\pgfsetroundjoin%
\pgfsetlinewidth{0.803000pt}%
\definecolor{currentstroke}{rgb}{0.690196,0.690196,0.690196}%
\pgfsetstrokecolor{currentstroke}%
\pgfsetdash{{2.960000pt}{1.280000pt}}{0.000000pt}%
\pgfpathmoveto{\pgfqpoint{0.961364in}{0.250000in}}%
\pgfpathlineto{\pgfqpoint{0.961364in}{1.760000in}}%
\pgfusepath{stroke}%
\end{pgfscope}%
\begin{pgfscope}%
\pgfsetbuttcap%
\pgfsetroundjoin%
\definecolor{currentfill}{rgb}{0.000000,0.000000,0.000000}%
\pgfsetfillcolor{currentfill}%
\pgfsetlinewidth{0.803000pt}%
\definecolor{currentstroke}{rgb}{0.000000,0.000000,0.000000}%
\pgfsetstrokecolor{currentstroke}%
\pgfsetdash{}{0pt}%
\pgfsys@defobject{currentmarker}{\pgfqpoint{0.000000in}{-0.048611in}}{\pgfqpoint{0.000000in}{0.000000in}}{%
\pgfpathmoveto{\pgfqpoint{0.000000in}{0.000000in}}%
\pgfpathlineto{\pgfqpoint{0.000000in}{-0.048611in}}%
\pgfusepath{stroke,fill}%
}%
\begin{pgfscope}%
\pgfsys@transformshift{0.961364in}{0.250000in}%
\pgfsys@useobject{currentmarker}{}%
\end{pgfscope}%
\end{pgfscope}%
\begin{pgfscope}%
\definecolor{textcolor}{rgb}{0.000000,0.000000,0.000000}%
\pgfsetstrokecolor{textcolor}%
\pgfsetfillcolor{textcolor}%
\pgftext[x=0.961364in,y=0.152778in,,top]{\color{textcolor}\rmfamily\fontsize{10.000000}{12.000000}\selectfont \(\displaystyle {3}\)}%
\end{pgfscope}%
\begin{pgfscope}%
\pgfpathrectangle{\pgfqpoint{0.750000in}{0.250000in}}{\pgfqpoint{4.650000in}{1.510000in}}%
\pgfusepath{clip}%
\pgfsetbuttcap%
\pgfsetroundjoin%
\pgfsetlinewidth{0.803000pt}%
\definecolor{currentstroke}{rgb}{0.690196,0.690196,0.690196}%
\pgfsetstrokecolor{currentstroke}%
\pgfsetdash{{2.960000pt}{1.280000pt}}{0.000000pt}%
\pgfpathmoveto{\pgfqpoint{2.370455in}{0.250000in}}%
\pgfpathlineto{\pgfqpoint{2.370455in}{1.760000in}}%
\pgfusepath{stroke}%
\end{pgfscope}%
\begin{pgfscope}%
\pgfsetbuttcap%
\pgfsetroundjoin%
\definecolor{currentfill}{rgb}{0.000000,0.000000,0.000000}%
\pgfsetfillcolor{currentfill}%
\pgfsetlinewidth{0.803000pt}%
\definecolor{currentstroke}{rgb}{0.000000,0.000000,0.000000}%
\pgfsetstrokecolor{currentstroke}%
\pgfsetdash{}{0pt}%
\pgfsys@defobject{currentmarker}{\pgfqpoint{0.000000in}{-0.048611in}}{\pgfqpoint{0.000000in}{0.000000in}}{%
\pgfpathmoveto{\pgfqpoint{0.000000in}{0.000000in}}%
\pgfpathlineto{\pgfqpoint{0.000000in}{-0.048611in}}%
\pgfusepath{stroke,fill}%
}%
\begin{pgfscope}%
\pgfsys@transformshift{2.370455in}{0.250000in}%
\pgfsys@useobject{currentmarker}{}%
\end{pgfscope}%
\end{pgfscope}%
\begin{pgfscope}%
\definecolor{textcolor}{rgb}{0.000000,0.000000,0.000000}%
\pgfsetstrokecolor{textcolor}%
\pgfsetfillcolor{textcolor}%
\pgftext[x=2.370455in,y=0.152778in,,top]{\color{textcolor}\rmfamily\fontsize{10.000000}{12.000000}\selectfont \(\displaystyle {4}\)}%
\end{pgfscope}%
\begin{pgfscope}%
\pgfpathrectangle{\pgfqpoint{0.750000in}{0.250000in}}{\pgfqpoint{4.650000in}{1.510000in}}%
\pgfusepath{clip}%
\pgfsetbuttcap%
\pgfsetroundjoin%
\pgfsetlinewidth{0.803000pt}%
\definecolor{currentstroke}{rgb}{0.690196,0.690196,0.690196}%
\pgfsetstrokecolor{currentstroke}%
\pgfsetdash{{2.960000pt}{1.280000pt}}{0.000000pt}%
\pgfpathmoveto{\pgfqpoint{3.779545in}{0.250000in}}%
\pgfpathlineto{\pgfqpoint{3.779545in}{1.760000in}}%
\pgfusepath{stroke}%
\end{pgfscope}%
\begin{pgfscope}%
\pgfsetbuttcap%
\pgfsetroundjoin%
\definecolor{currentfill}{rgb}{0.000000,0.000000,0.000000}%
\pgfsetfillcolor{currentfill}%
\pgfsetlinewidth{0.803000pt}%
\definecolor{currentstroke}{rgb}{0.000000,0.000000,0.000000}%
\pgfsetstrokecolor{currentstroke}%
\pgfsetdash{}{0pt}%
\pgfsys@defobject{currentmarker}{\pgfqpoint{0.000000in}{-0.048611in}}{\pgfqpoint{0.000000in}{0.000000in}}{%
\pgfpathmoveto{\pgfqpoint{0.000000in}{0.000000in}}%
\pgfpathlineto{\pgfqpoint{0.000000in}{-0.048611in}}%
\pgfusepath{stroke,fill}%
}%
\begin{pgfscope}%
\pgfsys@transformshift{3.779545in}{0.250000in}%
\pgfsys@useobject{currentmarker}{}%
\end{pgfscope}%
\end{pgfscope}%
\begin{pgfscope}%
\definecolor{textcolor}{rgb}{0.000000,0.000000,0.000000}%
\pgfsetstrokecolor{textcolor}%
\pgfsetfillcolor{textcolor}%
\pgftext[x=3.779545in,y=0.152778in,,top]{\color{textcolor}\rmfamily\fontsize{10.000000}{12.000000}\selectfont \(\displaystyle {5}\)}%
\end{pgfscope}%
\begin{pgfscope}%
\pgfpathrectangle{\pgfqpoint{0.750000in}{0.250000in}}{\pgfqpoint{4.650000in}{1.510000in}}%
\pgfusepath{clip}%
\pgfsetbuttcap%
\pgfsetroundjoin%
\pgfsetlinewidth{0.803000pt}%
\definecolor{currentstroke}{rgb}{0.690196,0.690196,0.690196}%
\pgfsetstrokecolor{currentstroke}%
\pgfsetdash{{2.960000pt}{1.280000pt}}{0.000000pt}%
\pgfpathmoveto{\pgfqpoint{5.188636in}{0.250000in}}%
\pgfpathlineto{\pgfqpoint{5.188636in}{1.760000in}}%
\pgfusepath{stroke}%
\end{pgfscope}%
\begin{pgfscope}%
\pgfsetbuttcap%
\pgfsetroundjoin%
\definecolor{currentfill}{rgb}{0.000000,0.000000,0.000000}%
\pgfsetfillcolor{currentfill}%
\pgfsetlinewidth{0.803000pt}%
\definecolor{currentstroke}{rgb}{0.000000,0.000000,0.000000}%
\pgfsetstrokecolor{currentstroke}%
\pgfsetdash{}{0pt}%
\pgfsys@defobject{currentmarker}{\pgfqpoint{0.000000in}{-0.048611in}}{\pgfqpoint{0.000000in}{0.000000in}}{%
\pgfpathmoveto{\pgfqpoint{0.000000in}{0.000000in}}%
\pgfpathlineto{\pgfqpoint{0.000000in}{-0.048611in}}%
\pgfusepath{stroke,fill}%
}%
\begin{pgfscope}%
\pgfsys@transformshift{5.188636in}{0.250000in}%
\pgfsys@useobject{currentmarker}{}%
\end{pgfscope}%
\end{pgfscope}%
\begin{pgfscope}%
\definecolor{textcolor}{rgb}{0.000000,0.000000,0.000000}%
\pgfsetstrokecolor{textcolor}%
\pgfsetfillcolor{textcolor}%
\pgftext[x=5.188636in,y=0.152778in,,top]{\color{textcolor}\rmfamily\fontsize{10.000000}{12.000000}\selectfont \(\displaystyle {6}\)}%
\end{pgfscope}%
\begin{pgfscope}%
\definecolor{textcolor}{rgb}{0.000000,0.000000,0.000000}%
\pgfsetstrokecolor{textcolor}%
\pgfsetfillcolor{textcolor}%
\pgftext[x=3.075000in,y=-0.026234in,,top]{\color{textcolor}\rmfamily\fontsize{10.000000}{12.000000}\selectfont c}%
\end{pgfscope}%
\begin{pgfscope}%
\pgfpathrectangle{\pgfqpoint{0.750000in}{0.250000in}}{\pgfqpoint{4.650000in}{1.510000in}}%
\pgfusepath{clip}%
\pgfsetbuttcap%
\pgfsetroundjoin%
\pgfsetlinewidth{0.803000pt}%
\definecolor{currentstroke}{rgb}{0.690196,0.690196,0.690196}%
\pgfsetstrokecolor{currentstroke}%
\pgfsetdash{{2.960000pt}{1.280000pt}}{0.000000pt}%
\pgfpathmoveto{\pgfqpoint{0.750000in}{0.311126in}}%
\pgfpathlineto{\pgfqpoint{5.400000in}{0.311126in}}%
\pgfusepath{stroke}%
\end{pgfscope}%
\begin{pgfscope}%
\pgfsetbuttcap%
\pgfsetroundjoin%
\definecolor{currentfill}{rgb}{0.000000,0.000000,0.000000}%
\pgfsetfillcolor{currentfill}%
\pgfsetlinewidth{0.803000pt}%
\definecolor{currentstroke}{rgb}{0.000000,0.000000,0.000000}%
\pgfsetstrokecolor{currentstroke}%
\pgfsetdash{}{0pt}%
\pgfsys@defobject{currentmarker}{\pgfqpoint{-0.048611in}{0.000000in}}{\pgfqpoint{-0.000000in}{0.000000in}}{%
\pgfpathmoveto{\pgfqpoint{-0.000000in}{0.000000in}}%
\pgfpathlineto{\pgfqpoint{-0.048611in}{0.000000in}}%
\pgfusepath{stroke,fill}%
}%
\begin{pgfscope}%
\pgfsys@transformshift{0.750000in}{0.311126in}%
\pgfsys@useobject{currentmarker}{}%
\end{pgfscope}%
\end{pgfscope}%
\begin{pgfscope}%
\definecolor{textcolor}{rgb}{0.000000,0.000000,0.000000}%
\pgfsetstrokecolor{textcolor}%
\pgfsetfillcolor{textcolor}%
\pgftext[x=0.513888in, y=0.262900in, left, base]{\color{textcolor}\rmfamily\fontsize{10.000000}{12.000000}\selectfont \(\displaystyle {75}\)}%
\end{pgfscope}%
\begin{pgfscope}%
\pgfpathrectangle{\pgfqpoint{0.750000in}{0.250000in}}{\pgfqpoint{4.650000in}{1.510000in}}%
\pgfusepath{clip}%
\pgfsetbuttcap%
\pgfsetroundjoin%
\pgfsetlinewidth{0.803000pt}%
\definecolor{currentstroke}{rgb}{0.690196,0.690196,0.690196}%
\pgfsetstrokecolor{currentstroke}%
\pgfsetdash{{2.960000pt}{1.280000pt}}{0.000000pt}%
\pgfpathmoveto{\pgfqpoint{0.750000in}{0.742468in}}%
\pgfpathlineto{\pgfqpoint{5.400000in}{0.742468in}}%
\pgfusepath{stroke}%
\end{pgfscope}%
\begin{pgfscope}%
\pgfsetbuttcap%
\pgfsetroundjoin%
\definecolor{currentfill}{rgb}{0.000000,0.000000,0.000000}%
\pgfsetfillcolor{currentfill}%
\pgfsetlinewidth{0.803000pt}%
\definecolor{currentstroke}{rgb}{0.000000,0.000000,0.000000}%
\pgfsetstrokecolor{currentstroke}%
\pgfsetdash{}{0pt}%
\pgfsys@defobject{currentmarker}{\pgfqpoint{-0.048611in}{0.000000in}}{\pgfqpoint{-0.000000in}{0.000000in}}{%
\pgfpathmoveto{\pgfqpoint{-0.000000in}{0.000000in}}%
\pgfpathlineto{\pgfqpoint{-0.048611in}{0.000000in}}%
\pgfusepath{stroke,fill}%
}%
\begin{pgfscope}%
\pgfsys@transformshift{0.750000in}{0.742468in}%
\pgfsys@useobject{currentmarker}{}%
\end{pgfscope}%
\end{pgfscope}%
\begin{pgfscope}%
\definecolor{textcolor}{rgb}{0.000000,0.000000,0.000000}%
\pgfsetstrokecolor{textcolor}%
\pgfsetfillcolor{textcolor}%
\pgftext[x=0.513888in, y=0.694242in, left, base]{\color{textcolor}\rmfamily\fontsize{10.000000}{12.000000}\selectfont \(\displaystyle {80}\)}%
\end{pgfscope}%
\begin{pgfscope}%
\pgfpathrectangle{\pgfqpoint{0.750000in}{0.250000in}}{\pgfqpoint{4.650000in}{1.510000in}}%
\pgfusepath{clip}%
\pgfsetbuttcap%
\pgfsetroundjoin%
\pgfsetlinewidth{0.803000pt}%
\definecolor{currentstroke}{rgb}{0.690196,0.690196,0.690196}%
\pgfsetstrokecolor{currentstroke}%
\pgfsetdash{{2.960000pt}{1.280000pt}}{0.000000pt}%
\pgfpathmoveto{\pgfqpoint{0.750000in}{1.173810in}}%
\pgfpathlineto{\pgfqpoint{5.400000in}{1.173810in}}%
\pgfusepath{stroke}%
\end{pgfscope}%
\begin{pgfscope}%
\pgfsetbuttcap%
\pgfsetroundjoin%
\definecolor{currentfill}{rgb}{0.000000,0.000000,0.000000}%
\pgfsetfillcolor{currentfill}%
\pgfsetlinewidth{0.803000pt}%
\definecolor{currentstroke}{rgb}{0.000000,0.000000,0.000000}%
\pgfsetstrokecolor{currentstroke}%
\pgfsetdash{}{0pt}%
\pgfsys@defobject{currentmarker}{\pgfqpoint{-0.048611in}{0.000000in}}{\pgfqpoint{-0.000000in}{0.000000in}}{%
\pgfpathmoveto{\pgfqpoint{-0.000000in}{0.000000in}}%
\pgfpathlineto{\pgfqpoint{-0.048611in}{0.000000in}}%
\pgfusepath{stroke,fill}%
}%
\begin{pgfscope}%
\pgfsys@transformshift{0.750000in}{1.173810in}%
\pgfsys@useobject{currentmarker}{}%
\end{pgfscope}%
\end{pgfscope}%
\begin{pgfscope}%
\definecolor{textcolor}{rgb}{0.000000,0.000000,0.000000}%
\pgfsetstrokecolor{textcolor}%
\pgfsetfillcolor{textcolor}%
\pgftext[x=0.513888in, y=1.125584in, left, base]{\color{textcolor}\rmfamily\fontsize{10.000000}{12.000000}\selectfont \(\displaystyle {85}\)}%
\end{pgfscope}%
\begin{pgfscope}%
\pgfpathrectangle{\pgfqpoint{0.750000in}{0.250000in}}{\pgfqpoint{4.650000in}{1.510000in}}%
\pgfusepath{clip}%
\pgfsetbuttcap%
\pgfsetroundjoin%
\pgfsetlinewidth{0.803000pt}%
\definecolor{currentstroke}{rgb}{0.690196,0.690196,0.690196}%
\pgfsetstrokecolor{currentstroke}%
\pgfsetdash{{2.960000pt}{1.280000pt}}{0.000000pt}%
\pgfpathmoveto{\pgfqpoint{0.750000in}{1.605152in}}%
\pgfpathlineto{\pgfqpoint{5.400000in}{1.605152in}}%
\pgfusepath{stroke}%
\end{pgfscope}%
\begin{pgfscope}%
\pgfsetbuttcap%
\pgfsetroundjoin%
\definecolor{currentfill}{rgb}{0.000000,0.000000,0.000000}%
\pgfsetfillcolor{currentfill}%
\pgfsetlinewidth{0.803000pt}%
\definecolor{currentstroke}{rgb}{0.000000,0.000000,0.000000}%
\pgfsetstrokecolor{currentstroke}%
\pgfsetdash{}{0pt}%
\pgfsys@defobject{currentmarker}{\pgfqpoint{-0.048611in}{0.000000in}}{\pgfqpoint{-0.000000in}{0.000000in}}{%
\pgfpathmoveto{\pgfqpoint{-0.000000in}{0.000000in}}%
\pgfpathlineto{\pgfqpoint{-0.048611in}{0.000000in}}%
\pgfusepath{stroke,fill}%
}%
\begin{pgfscope}%
\pgfsys@transformshift{0.750000in}{1.605152in}%
\pgfsys@useobject{currentmarker}{}%
\end{pgfscope}%
\end{pgfscope}%
\begin{pgfscope}%
\definecolor{textcolor}{rgb}{0.000000,0.000000,0.000000}%
\pgfsetstrokecolor{textcolor}%
\pgfsetfillcolor{textcolor}%
\pgftext[x=0.513888in, y=1.556926in, left, base]{\color{textcolor}\rmfamily\fontsize{10.000000}{12.000000}\selectfont \(\displaystyle {90}\)}%
\end{pgfscope}%
\begin{pgfscope}%
\definecolor{textcolor}{rgb}{0.000000,0.000000,0.000000}%
\pgfsetstrokecolor{textcolor}%
\pgfsetfillcolor{textcolor}%
\pgftext[x=0.458333in,y=1.005000in,,bottom,rotate=90.000000]{\color{textcolor}\rmfamily\fontsize{10.000000}{12.000000}\selectfont AUC}%
\end{pgfscope}%
\begin{pgfscope}%
\pgfpathrectangle{\pgfqpoint{0.750000in}{0.250000in}}{\pgfqpoint{4.650000in}{1.510000in}}%
\pgfusepath{clip}%
\pgfsetrectcap%
\pgfsetroundjoin%
\pgfsetlinewidth{1.505625pt}%
\definecolor{currentstroke}{rgb}{0.121569,0.466667,0.705882}%
\pgfsetstrokecolor{currentstroke}%
\pgfsetdash{}{0pt}%
\pgfpathmoveto{\pgfqpoint{0.961364in}{0.318636in}}%
\pgfpathlineto{\pgfqpoint{0.961364in}{1.276266in}}%
\pgfusepath{stroke}%
\end{pgfscope}%
\begin{pgfscope}%
\pgfpathrectangle{\pgfqpoint{0.750000in}{0.250000in}}{\pgfqpoint{4.650000in}{1.510000in}}%
\pgfusepath{clip}%
\pgfsetrectcap%
\pgfsetroundjoin%
\pgfsetlinewidth{1.505625pt}%
\definecolor{currentstroke}{rgb}{0.121569,0.466667,0.705882}%
\pgfsetstrokecolor{currentstroke}%
\pgfsetdash{}{0pt}%
\pgfpathmoveto{\pgfqpoint{2.370455in}{0.769568in}}%
\pgfpathlineto{\pgfqpoint{2.370455in}{1.691364in}}%
\pgfusepath{stroke}%
\end{pgfscope}%
\begin{pgfscope}%
\pgfpathrectangle{\pgfqpoint{0.750000in}{0.250000in}}{\pgfqpoint{4.650000in}{1.510000in}}%
\pgfusepath{clip}%
\pgfsetrectcap%
\pgfsetroundjoin%
\pgfsetlinewidth{1.505625pt}%
\definecolor{currentstroke}{rgb}{0.121569,0.466667,0.705882}%
\pgfsetstrokecolor{currentstroke}%
\pgfsetdash{}{0pt}%
\pgfpathmoveto{\pgfqpoint{3.779545in}{0.447446in}}%
\pgfpathlineto{\pgfqpoint{3.779545in}{1.141049in}}%
\pgfusepath{stroke}%
\end{pgfscope}%
\begin{pgfscope}%
\pgfpathrectangle{\pgfqpoint{0.750000in}{0.250000in}}{\pgfqpoint{4.650000in}{1.510000in}}%
\pgfusepath{clip}%
\pgfsetrectcap%
\pgfsetroundjoin%
\pgfsetlinewidth{1.505625pt}%
\definecolor{currentstroke}{rgb}{0.121569,0.466667,0.705882}%
\pgfsetstrokecolor{currentstroke}%
\pgfsetdash{}{0pt}%
\pgfpathmoveto{\pgfqpoint{5.188636in}{0.409480in}}%
\pgfpathlineto{\pgfqpoint{5.188636in}{1.069452in}}%
\pgfusepath{stroke}%
\end{pgfscope}%
\begin{pgfscope}%
\pgfpathrectangle{\pgfqpoint{0.750000in}{0.250000in}}{\pgfqpoint{4.650000in}{1.510000in}}%
\pgfusepath{clip}%
\pgfsetrectcap%
\pgfsetroundjoin%
\pgfsetlinewidth{1.505625pt}%
\definecolor{currentstroke}{rgb}{0.121569,0.466667,0.705882}%
\pgfsetstrokecolor{currentstroke}%
\pgfsetdash{}{0pt}%
\pgfpathmoveto{\pgfqpoint{0.961364in}{0.797451in}}%
\pgfpathlineto{\pgfqpoint{2.370455in}{1.230466in}}%
\pgfpathlineto{\pgfqpoint{3.779545in}{0.794247in}}%
\pgfpathlineto{\pgfqpoint{5.188636in}{0.739466in}}%
\pgfusepath{stroke}%
\end{pgfscope}%
\begin{pgfscope}%
\pgfpathrectangle{\pgfqpoint{0.750000in}{0.250000in}}{\pgfqpoint{4.650000in}{1.510000in}}%
\pgfusepath{clip}%
\pgfsetbuttcap%
\pgfsetroundjoin%
\definecolor{currentfill}{rgb}{0.121569,0.466667,0.705882}%
\pgfsetfillcolor{currentfill}%
\pgfsetlinewidth{1.003750pt}%
\definecolor{currentstroke}{rgb}{0.121569,0.466667,0.705882}%
\pgfsetstrokecolor{currentstroke}%
\pgfsetdash{}{0pt}%
\pgfsys@defobject{currentmarker}{\pgfqpoint{-0.055556in}{-0.000000in}}{\pgfqpoint{0.055556in}{0.000000in}}{%
\pgfpathmoveto{\pgfqpoint{0.055556in}{-0.000000in}}%
\pgfpathlineto{\pgfqpoint{-0.055556in}{0.000000in}}%
\pgfusepath{stroke,fill}%
}%
\begin{pgfscope}%
\pgfsys@transformshift{0.961364in}{0.318636in}%
\pgfsys@useobject{currentmarker}{}%
\end{pgfscope}%
\begin{pgfscope}%
\pgfsys@transformshift{2.370455in}{0.769568in}%
\pgfsys@useobject{currentmarker}{}%
\end{pgfscope}%
\begin{pgfscope}%
\pgfsys@transformshift{3.779545in}{0.447446in}%
\pgfsys@useobject{currentmarker}{}%
\end{pgfscope}%
\begin{pgfscope}%
\pgfsys@transformshift{5.188636in}{0.409480in}%
\pgfsys@useobject{currentmarker}{}%
\end{pgfscope}%
\end{pgfscope}%
\begin{pgfscope}%
\pgfpathrectangle{\pgfqpoint{0.750000in}{0.250000in}}{\pgfqpoint{4.650000in}{1.510000in}}%
\pgfusepath{clip}%
\pgfsetbuttcap%
\pgfsetroundjoin%
\definecolor{currentfill}{rgb}{0.121569,0.466667,0.705882}%
\pgfsetfillcolor{currentfill}%
\pgfsetlinewidth{1.003750pt}%
\definecolor{currentstroke}{rgb}{0.121569,0.466667,0.705882}%
\pgfsetstrokecolor{currentstroke}%
\pgfsetdash{}{0pt}%
\pgfsys@defobject{currentmarker}{\pgfqpoint{-0.055556in}{-0.000000in}}{\pgfqpoint{0.055556in}{0.000000in}}{%
\pgfpathmoveto{\pgfqpoint{0.055556in}{-0.000000in}}%
\pgfpathlineto{\pgfqpoint{-0.055556in}{0.000000in}}%
\pgfusepath{stroke,fill}%
}%
\begin{pgfscope}%
\pgfsys@transformshift{0.961364in}{1.276266in}%
\pgfsys@useobject{currentmarker}{}%
\end{pgfscope}%
\begin{pgfscope}%
\pgfsys@transformshift{2.370455in}{1.691364in}%
\pgfsys@useobject{currentmarker}{}%
\end{pgfscope}%
\begin{pgfscope}%
\pgfsys@transformshift{3.779545in}{1.141049in}%
\pgfsys@useobject{currentmarker}{}%
\end{pgfscope}%
\begin{pgfscope}%
\pgfsys@transformshift{5.188636in}{1.069452in}%
\pgfsys@useobject{currentmarker}{}%
\end{pgfscope}%
\end{pgfscope}%
\begin{pgfscope}%
\pgfsetrectcap%
\pgfsetmiterjoin%
\pgfsetlinewidth{0.803000pt}%
\definecolor{currentstroke}{rgb}{0.000000,0.000000,0.000000}%
\pgfsetstrokecolor{currentstroke}%
\pgfsetdash{}{0pt}%
\pgfpathmoveto{\pgfqpoint{0.750000in}{0.250000in}}%
\pgfpathlineto{\pgfqpoint{0.750000in}{1.760000in}}%
\pgfusepath{stroke}%
\end{pgfscope}%
\begin{pgfscope}%
\pgfsetrectcap%
\pgfsetmiterjoin%
\pgfsetlinewidth{0.803000pt}%
\definecolor{currentstroke}{rgb}{0.000000,0.000000,0.000000}%
\pgfsetstrokecolor{currentstroke}%
\pgfsetdash{}{0pt}%
\pgfpathmoveto{\pgfqpoint{5.400000in}{0.250000in}}%
\pgfpathlineto{\pgfqpoint{5.400000in}{1.760000in}}%
\pgfusepath{stroke}%
\end{pgfscope}%
\begin{pgfscope}%
\pgfsetrectcap%
\pgfsetmiterjoin%
\pgfsetlinewidth{0.803000pt}%
\definecolor{currentstroke}{rgb}{0.000000,0.000000,0.000000}%
\pgfsetstrokecolor{currentstroke}%
\pgfsetdash{}{0pt}%
\pgfpathmoveto{\pgfqpoint{0.750000in}{0.250000in}}%
\pgfpathlineto{\pgfqpoint{5.400000in}{0.250000in}}%
\pgfusepath{stroke}%
\end{pgfscope}%
\begin{pgfscope}%
\pgfsetrectcap%
\pgfsetmiterjoin%
\pgfsetlinewidth{0.803000pt}%
\definecolor{currentstroke}{rgb}{0.000000,0.000000,0.000000}%
\pgfsetstrokecolor{currentstroke}%
\pgfsetdash{}{0pt}%
\pgfpathmoveto{\pgfqpoint{0.750000in}{1.760000in}}%
\pgfpathlineto{\pgfqpoint{5.400000in}{1.760000in}}%
\pgfusepath{stroke}%
\end{pgfscope}%
\end{pgfpicture}%
\makeatother%
\endgroup%